\documentclass{article}  
\usepackage{iclr2026_conference,times}

\usepackage{amsmath,amsfonts,bm}

\def\eqref#1{equation~\ref{#1}}

\def\1{\bm{1}}

\DeclareMathAlphabet{\mathsfit}{\encodingdefault}{\sfdefault}{m}{sl}
\SetMathAlphabet{\mathsfit}{bold}{\encodingdefault}{\sfdefault}{bx}{n}

\usepackage{capt-of, eucal}
\usepackage{xspace}
\usepackage{xcolor}
\usepackage{enumitem}
\usepackage{amsmath,amsthm,amssymb,amsfonts,dsfont,pifont,bm,bbm,mathrsfs,mathtools,nicefrac,extarrows,relsize}
\usepackage{algorithm,algpseudocode,listings}
\usepackage{booktabs,multirow,adjustbox,diagbox,threeparttable,tabularray,setspace}
\usepackage{wrapfig}
\usepackage{graphicx}
\usepackage{subcaption}
\usepackage{tabularx}
\usepackage{array}
\usepackage{makecell}
\usepackage{longtable} 
\usepackage{paracol}
\usepackage{float}
\usepackage{fontawesome5}
\usepackage[tableposition=top]{caption}
\usepackage{colortbl}
\usepackage{color}
\usepackage{multicol}
\usepackage{lipsum}

\definecolor{citeblue}{rgb}{0.21,0.49,0.74}
\usepackage[pagebackref=false,breaklinks,colorlinks,citecolor=citeblue,bookmarks=false]{hyperref}
\usepackage{wrapfig}
\usepackage[capitalize]{cleveref}  

\crefname{section}{Sec.}{Secs.}
\Crefname{section}{Section}{Sections}
\crefname{appendix}{Appendix}{Appendices}
\Crefname{appendix}{Appendix}{Appendices}
\crefname{table}{Table}{Tables}
\Crefname{table}{Table}{Tables}
\crefname{figure}{Fig.}{Figs.}
\Crefname{figure}{Figure}{Figures}
\crefname{equation}{Eq.}{Eqs.}
\Crefname{equation}{Equation}{Equations}
\crefname{theorem}{Thm.}{Thms.}
\Crefname{theorem}{Theorem}{Theorems}
\crefname{lemma}{Lem.}{Lems.}
\Crefname{lemma}{Lemma}{Lemmas}
\crefname{remark}{Rem.}{Rems.}
\Crefname{remark}{Remark}{Remarks}
\crefname{corollary}{Cor.}{Cors.}
\Crefname{corollary}{Corollary}{Corollaries}
\crefname{algorithm}{Alg.}{Algs.}
\Crefname{algorithm}{Algorithm}{Algorithms}
\definecolor{ourreward}{HTML}{32bbee}
\definecolor{sparsereward}{HTML}{ed9a81}
\definecolor{cellred}{RGB}{213, 123, 101}
\definecolor{cellgreen}{RGB}{0, 205, 0}
\definecolor{cellblue}{RGB}{54, 125, 189}
\definecolor{codegreen}{rgb}{0,0.6,0}
\definecolor{codegray}{rgb}{0.5,0.5,0.5}
\definecolor{codepurple}{rgb}{0.58,0,0.82}
\definecolor{backcolour}{rgb}{1.0,1.0,1.0}
\lstdefinestyle{mystyle}{
    backgroundcolor=\color{backcolour},
    commentstyle=\color{codegreen},
    keywordstyle=\color{magenta},
    numberstyle=\tiny\color{codegray},
    stringstyle=\color{codepurple},
    basicstyle=\ttfamily\scriptsize,
    breakatwhitespace=false,
    breaklines=true,
    captionpos=b,
    keepspaces=true,
    numbers=left,
    numbersep=5pt,
    showspaces=false,
    showstringspaces=false,
    showtabs=false,
    tabsize=2
}
\usepackage[most,skins,theorems]{tcolorbox}
\tcbset{
  aibox/.style={
    width=\linewidth,
    top=8pt,
    bottom=4pt,
    colback=blue!6!white,
    colframe=black,
    colbacktitle=black,
    enhanced,
    center,
    attach boxed title to top left={yshift=-0.1in,xshift=0.15in},
    boxed title style={boxrule=0pt,colframe=white,},
  }
}
\newtcolorbox{AIbox}[2][]{aibox,title=#2,#1}
\usepackage{amsmath} 
\usepackage{amssymb} 
\usepackage{multirow}

\usepackage{rotating}  

\newcolumntype{C}[1]{>{\centering\arraybackslash}p{#1}}
\newcolumntype{L}[1]{>{\arraybackslash}p{#1}}

\definecolor{demphcolor}{gray}{.2}

\definecolor{demphcolorinline}{gray}{.3}

\definecolor{demphcolor1}{gray}{.6}

\newcommand{\tocite}[1]{{\color{red} [TO CITE]}}

\title{%
  \texorpdfstring{%
    \raisebox{-0.23\height}{\includegraphics[scale=0.05]{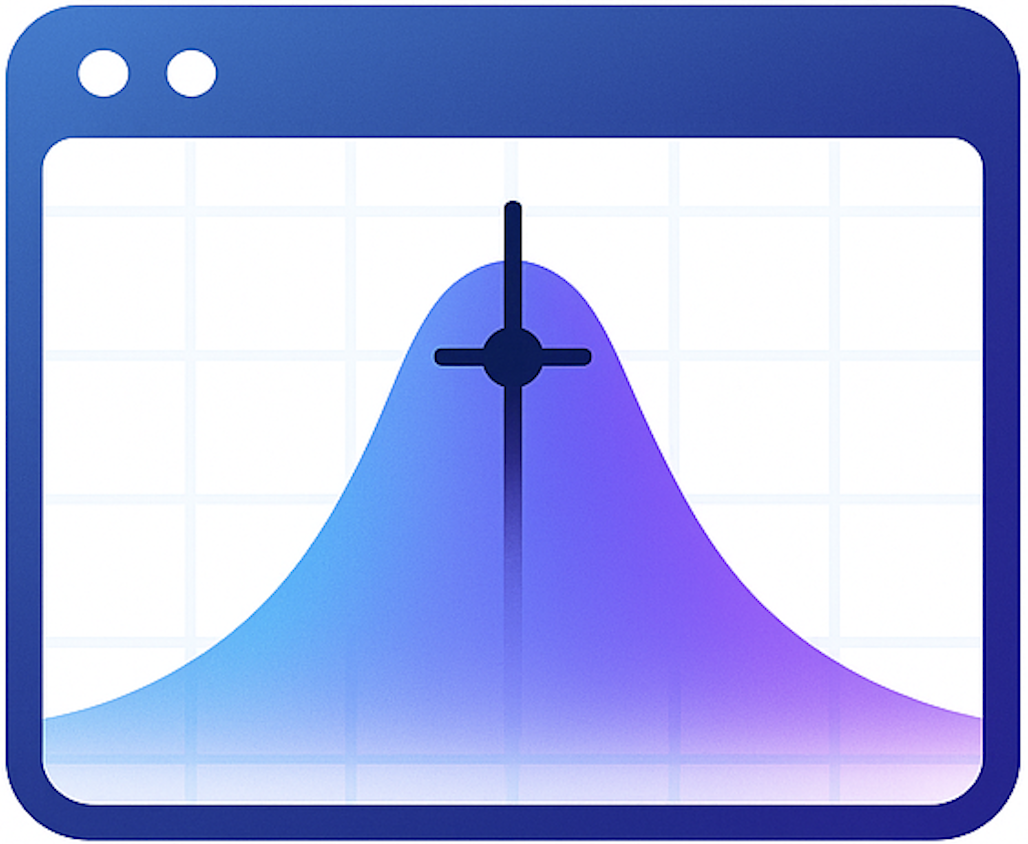}}%
    ~GUI-G$^2$: Gaussian Reward Modeling\\for GUI Grounding%
  }{GUI-G$^2$: Gaussian Reward Modeling\\for GUI Grounding}%
}

\newcommand{\methodname}{GUI-G$^2$\xspace}

\author{%
  \textbf{Fei Tang}$^{1,2}$\thanks{~This work was done when the first author was an intern at Ant Group.}, 
  ~~
  \textbf{Zhangxuan Gu}$^{2}$, 
  ~~
  \textbf{Zhengxi Lu}$^{1}$, 
  ~~
  \textbf{Xuyang Liu}$^{2}$\\
  \textbf{Shuheng Shen}$^{2}$, 
  ~~
  \textbf{Changhua Meng}$^{2}$, 
  ~~ 
  \textbf{Wen Wang}$^{1}$, 
  ~~ 
  \textbf{Wenqi Zhang}$^{1}$ \\
  \textbf{Yongliang Shen}$^{1}\thanks{~Corresponding author.}$, 
  ~~ 
  \textbf{Weiming Lu}$^{1}$, 
  ~~ 
  \textbf{Jun Xiao}$^{1}$, 
  ~~
  \textbf{Yueting Zhuang}$^{1}$ \\
  \vspace{-6pt}\\
  $^1$Zhejiang University, 
  ~~ 
  $^2$Ant Group \\
  \texttt{\{flysugar, syl\}@zju.edu.cn} \quad \texttt{shuheng.ssh@antgroup.com} \\
  \vspace{-6pt}\\
  \begin{tabular}{@{}ll@{}}
    \faGithub\ GitHub: & \href{https://github.com/zju-real/GUI-G2}{\texttt{\textcolor{cyan}{https://github.com/zju-real/GUI-G2}}} \\
    \faGlobe\ Project: & \href{https://zju-real.github.io/GUI-G2/}{\texttt{\textcolor{cyan}{https://zju-real.github.io/GUI-G2}}}
  \end{tabular}
}

\iclrfinalcopy 

\begin{document}

\maketitle

\vspace{-10mm}
\begin{figure}[htbp]
    \centering
    \begin{subfigure}{0.57\textwidth}
        \centering
        \includegraphics[width=\textwidth]{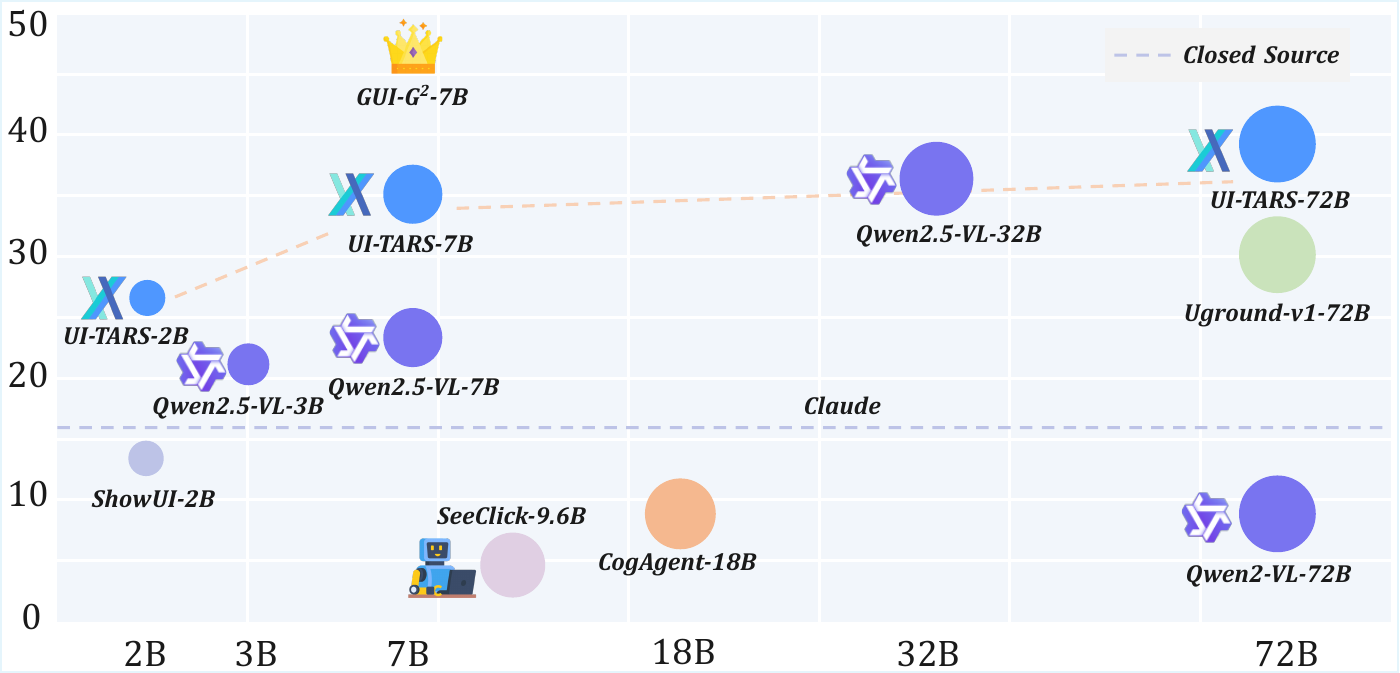}
    \end{subfigure}
    \hfill
    \begin{subfigure}{0.42\textwidth}
        \centering
        \includegraphics[width=\textwidth]{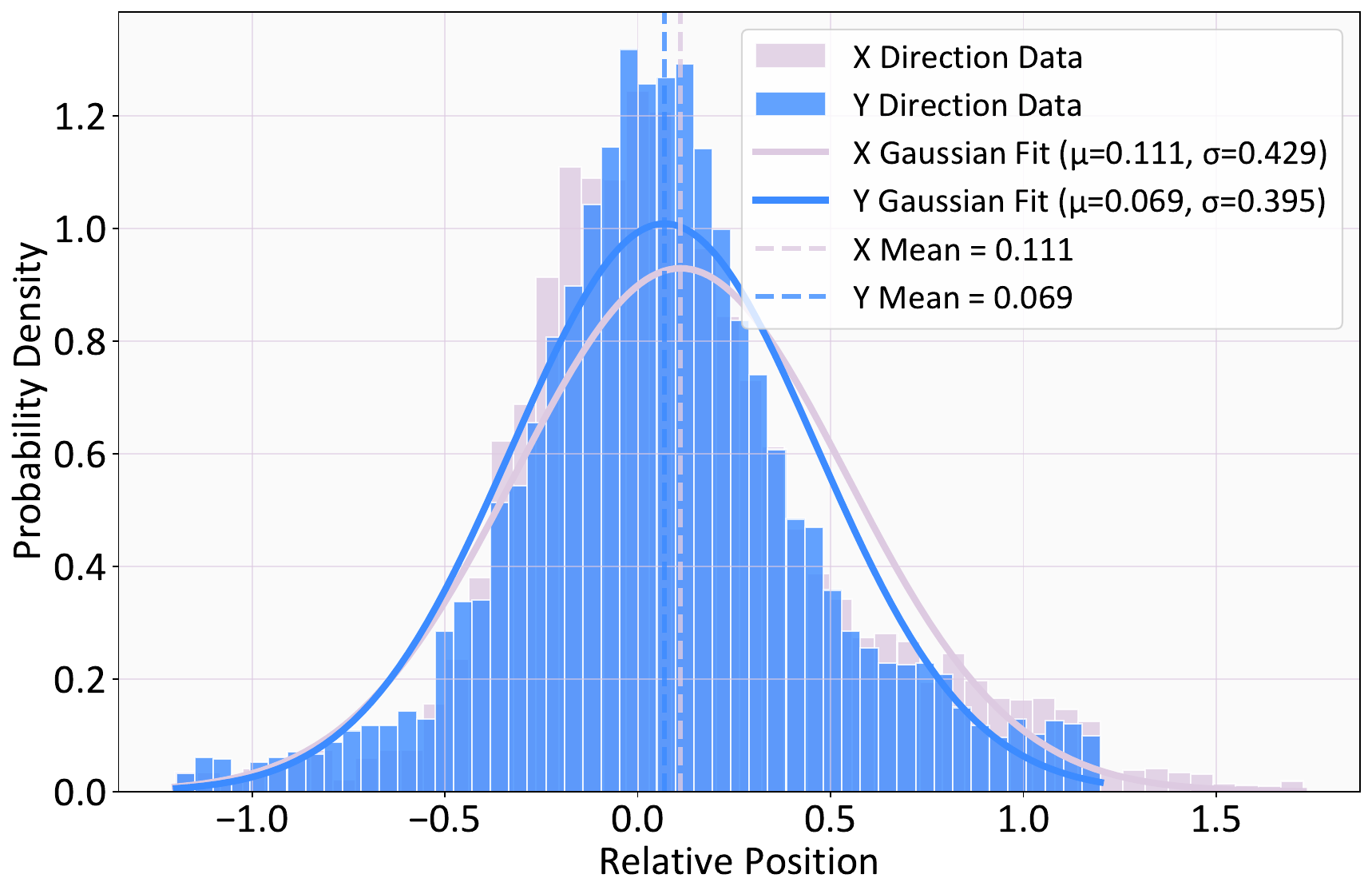}
         \vspace{-6mm}
    \end{subfigure}
    \caption{GUI grounding performance and human click behavior. Left: Performance comparison of various models on ScreenSpot-Pro. Right: Human click distribution from AITW~\citep{rawles2023androidwildlargescaledataset} reveals natural Gaussian patterns around target centers ($\mu=0.111$, $\sigma=0.429$), validating our design choice of continuous Gaussian rewards over discrete binary feedback.}
    \label{fig:leida}
\end{figure}

\begin{abstract}
Graphical User Interface (GUI) grounding maps natural language instructions to precise interface locations for autonomous interaction.
Current reinforcement learning approaches use binary rewards that treat elements as hit-or-miss targets, creating sparse signals that ignore the continuous nature of spatial interactions.
Motivated by human clicking behavior that naturally forms Gaussian distributions centered on target elements, we introduce GUI Gaussian Grounding Rewards (\methodname), a principled reward framework that models GUI elements as continuous Gaussian distributions across the interface plane.
\methodname incorporates two synergistic mechanisms: Gaussian point rewards model precise localization through exponentially decaying distributions centered on element centroids, while coverage rewards assess spatial alignment by measuring the overlap between predicted Gaussian distributions and target regions.
To handle diverse element scales, we develop an adaptive variance mechanism that calibrates reward distributions based on element dimensions.
This framework transforms GUI grounding from sparse binary classification to dense continuous optimization, where Gaussian distributions generate rich gradient signals that guide models toward optimal interaction positions.
Extensive experiments across ScreenSpot, ScreenSpot-v2, and ScreenSpot-Pro benchmarks demonstrate that \methodname, substantially outperforms state-of-the-art method UI-TARS-72B, with the most significant improvement of 24.7\% on ScreenSpot-Pro. Our analysis reveals that continuous modeling provides superior robustness to interface variations and enhanced generalization to unseen layouts, establishing a new paradigm for spatial reasoning in GUI interaction tasks.

\end{abstract}

\begin{figure}[t]
    \centering
    \includegraphics[width=1.0\textwidth]{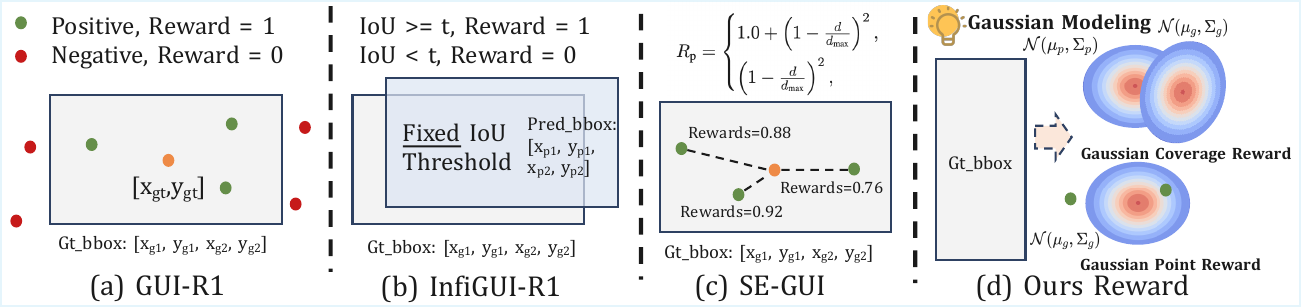}
    \caption{Comparison of reward modeling strategies. (a-c) Existing methods treat GUI elements as abstract points with binary or distance-based rewards, while (d) our Gaussian approach provides continuous point and coverage rewards that naturally align with human clicking behavior.}
    \label{fig:reward_com}
\end{figure}

\section{Introduction}
Autonomous GUI agents are revolutionizing human-computer interaction by allowing users to control interfaces with natural language across various applications~\citep{gou2024navigatingdigitalworldhumans,tang2025surveymllmbasedguiagents,cheng2024seeclickharnessingguigrounding}.
As the core of these systems, GUI grounding, is the fundamental capability to accurately map natural language instructions to precise pixel coordinates on interface elements~\citep{tang2025thinktwiceclickonce,cheng2024seeclickharnessingguigrounding,lin2024showuivisionlanguageactionmodelgui,guiactor}.

Recent advances in GUI grounding have increasingly adopted reinforcement learning frameworks~\citep{lu2025uir1enhancingefficientaction,luo2025guir1generalistr1style,liu2025infiguir1advancingmultimodalgui}. However, current approaches rely on binary reward systems~\citep{lu2025uir1enhancingefficientaction,luo2025guir1generalistr1style,yuan2025enhancingvisualgroundinggui,zhou2025guig1understandingr1zeroliketraining} that assign rewards of 1 for coordinates within target bounding boxes and 0 otherwise. This formulation treats GUI interactions as binary hit-or-miss problems, creating sparse learning signals where predictions one pixel outside target regions receive the same zero reward as complete failures (Figure~\ref{fig:reward_com}, a-b). The binary paradigm ignores two critical aspects of interface interaction: first, clicking quality varies continuously with distance from element centers, and second, interface elements are inherently two-dimensional regions with spatial structure, not abstract points (Figure~\ref{fig:reward_com}, a-c). This mismatch between discrete optimization and the continuous geometric nature of GUI interactions severely limits learning efficiency, particularly during early training when models need dense feedback to develop appropriate grounding behaviors.

This discrete approach contradicts empirical evidence from human behavior. Analysis of the AITW dataset~\citep{rawles2023androidwildlargescaledataset} reveals that users' clicks naturally form Gaussian distributions centered on target elements (Figure \ref{fig:framework}, right), consistent with Fitts' Law~\citep{fitts1954information,mackenzie1992fitts}. This pattern demonstrates that spatial targeting inherently follows continuous probability distributions, with click density decreasing smoothly from element centers to edges. Current binary mechanisms completely ignore this fundamental characteristic of human-computer interaction.

Building on this insight, we introduce \methodname (GUI Gaussian Grounding Rewards), a principled framework that fundamentally reconceptualizes GUI grounding by modeling clicking points as smooth probability distributions across the interface plane. Rather than treating elements as discrete hit-or-miss targets, \methodname represents them as continuous Gaussian distributions that provide rich spatial information and dense learning signals. This approach comprises two complementary mechanisms: First, we design point-based rewards that decrease smoothly with distance from element centers, encouraging precise localization while maintaining continuous gradients. Second, we introduce coverage-based rewards that measure the spatial overlap between predicted click distributions and target element regions, ensuring comprehensive element targeting.

To accommodate varying element scales, we introduce an adaptive variance mechanism that dynamically adjusts reward distributions according to element dimensions. This ensures consistent learning signals across GUI components while maintaining their distinct geometric properties. \methodname transforms GUI grounding from sparse binary optimization to dense continuous reasoning, enabling models to learn fine-grained spatial relationships and develop more robust interaction strategies.

Extensive evaluation on ScreenSpot~\citep{cheng2024seeclickharnessingguigrounding}, ScreenSpot-v2~\citep{wu2024osatlasfoundationactionmodel}, and ScreenSpot-Pro~\citep{li2024screenspot-pro} benchmarks demonstrates that our approach achieves substantial improvements over state-of-the-art methods, with accuracy gains up to 4.1\%, 3.3\%, and 24.7\% respectively. Our analysis reveals superior robustness to interface variations and enhanced generalization to unseen layouts, confirming that continuous spatial modeling provides more fundamental and transferable representations than discrete alternatives. Comprehensive ablation studies validate the synergistic contributions of both Gaussian components and the critical importance of adaptive variance mechanisms for handling interface diversity.

Our contributions are threefold:
\begin{itemize}
\item We introduce \methodname, a principled approach that models GUI interactions as continuous spatial processes, fundamentally transforming reward design from discrete binary signals to geometrically-aware continuous feedback that captures the inherent planar nature of interface elements.

\item We propose a novel dual-component reward system comprising Gaussian point rewards for precise localization and Gaussian coverage rewards for regional assessment, enhanced with adaptive variance mechanisms that automatically calibrate distributions based on element dimensions.

\item We demonstrate through extensive experiments that \methodname achieves substantial improvements, with accuracy of 92.0\% on ScreenSpot, 93.3\% on ScreenSpot-v2, and 47.5\% on ScreenSpot-Pro, while exhibiting superior robustness and generalization compared to discrete reward approaches.
\end{itemize}

\section{Related Work}
\subsection{GUI Agents}
GUI agents are intelligent systems that can understand and interact with graphical user interfaces through natural language instructions, enabling automated execution of complex computer tasks~\citep{gou2024navigatingdigitalworldhumans,zhang2025largelanguagemodelbrainedgui,tang2025surveymllmbasedguiagents,sun2025osgenesisautomatingguiagent,shen2023hugginggptsolvingaitasks,hong2024cogagentvisuallanguagemodel,yang2024ariauivisualgroundinggui}. These approaches can be broadly categorized into two main paradigms: 
\textit{\textbf{(1) Expert Design-Driven Workflow Paradigm}}: These approaches typically leverage closed-source multimodal large language models and construct workflows through expertly designed fine-grained modules such as planners~\citep{wang2024mobileagentautonomousmultimodalmobile,zhang2024ufouifocusedagentwindows} and grounders~\citep{gou2024navigatingdigitalworldhumans,liu2024autoglmautonomousfoundationagents,lin2024showuivisionlanguageactionmodelgui,wu2024osatlasfoundationactionmodel}. The Mobile-Agent series~\citep{wang2025mobileagenteselfevolvingmobileassistant,wang2024mobileagentv2mobiledeviceoperation,wang2024mobileagentautonomousmultimodalmobile}, AppAgent series~\citep{zhang2023appagentmultimodalagentssmartphone,li2024appagentv2advancedagent,jiang2025appagentxevolvingguiagents,xie2025scalingcomputerusegroundinguser}, and UFO series~\citep{zhang2024ufouifocusedagentwindows,zhang2025ufo2desktopagentos} all accomplish various tasks through these workflow-based approaches. These GUI agents typically consist of planners and grounders, where planners usually employ closed-source large language models such as GPT-4o~\citep{openai2024gpt4o} and Claude~\citep{anthropic2024cuda} for task planning. For grounding components, there are two main approaches: one utilizes HTML and DOM tree structures for screen understanding~\citep{rawles2023androidwildlargescaledataset,zhang2023appagentmultimodalagentssmartphone}, while the other employs visual tools such as OCR~\citep{du2020ppocrpracticalultralightweight}, SAM~\citep{kirillov2023segment}, and Omniparser~\citep{lu2024omniparserpurevisionbased} for more effective screen understanding and element localization. However, this reliance on pre-programmed workflows, driven by human expertise, makes frameworks inherently non-scalable, consuming substantial manual effort and proving difficult to extend to new domains~\citep{qin2025uitarspioneeringautomatedgui}.
\textit{\textbf{(2) Data-Driven Training Paradigm}}: These approaches employ specialized MLLMs trained specifically for GUI understanding and interaction through data-driven methodologies~\citep{qin2025uitarspioneeringautomatedgui,gou2024navigatingdigitalworldhumans,lin2024showuivisionlanguageactionmodelgui,cheng2024seeclickharnessingguigrounding,tang2025thinktwiceclickonce,wu2024osatlasfoundationactionmodel}. These works achieve GUI-specific capabilities by collecting large-scale GUI corpora for fine-tuning to develop models tailored for GUI tasks. For example, UI-TARS~\citep{qin2025uitarspioneeringautomatedgui} develops an end-to-end native GUI agent through large-scale GUI screenshots for enhanced perception and action traces for unified action modeling across platforms. However, due to the limitations of supervised fine-tuning~\citep{chu2025sftmemorizesrlgeneralizes}, these methods still face generalization challenges when encountering novel interface scenarios~\citep{luo2025guir1generalistr1style,lu2025uir1enhancingefficientaction}.
\subsection{Reinforcement Fine-Tuning}
Since the release of DeepSeek-R1~\citep{deepseekai2025deepseekr1incentivizingreasoningcapability}, rule-based reward reinforcement learning has been applied across various domains, such as video understanding~\citep{feng2025videor1reinforcingvideoreasoning} and multimodal reasoning~\citep{shen2025vlmr1stablegeneralizabler1style}. Researchers have begun applying this approach to GUI tasks. GUI-R1~\citep{luo2025guir1generalistr1style} and UI-R1~\citep{lu2025uir1enhancingefficientaction} apply verifiable reward paradigms to GUI tasks, representing pioneering efforts in this direction while demonstrating the potential of RFT. InfiGUI-R1~\citep{liu2025infiguir1advancingmultimodalgui} similarly follows the R1 paradigm, employing two-stage training to inject reasoning capabilities into the model. GUI-G1~\citep{zhou2025guig1understandingr1zeroliketraining} reanalyzes existing problems in current R1-based GUI agents and designs controllable box size rewards for GUI grounding tasks, while incorporating difficulty coefficient factors based on box size using the GRPO~\citep{shao2024deepseekmathpushinglimitsmathematical} algorithm to enable better learning. SE-GUI~\citep{yuan2025enhancingvisualgroundinggui} proposes self-evolution approaches and continuous rewards to guide model learning. 
However, most previous methods treat GUI elements as discrete point requiring perfect targeting and provide only sparse hit-or-miss feedback, struggling to provide effective guidance for model learning during the early stages of training. We address the limitations by proposing a Gaussian continuous reward mechanism, which provides dense and informative feedback to guide model learning more effectively.

\section{Method}

\begin{figure}[t]
    \centering
    \includegraphics[width=1.0\textwidth]{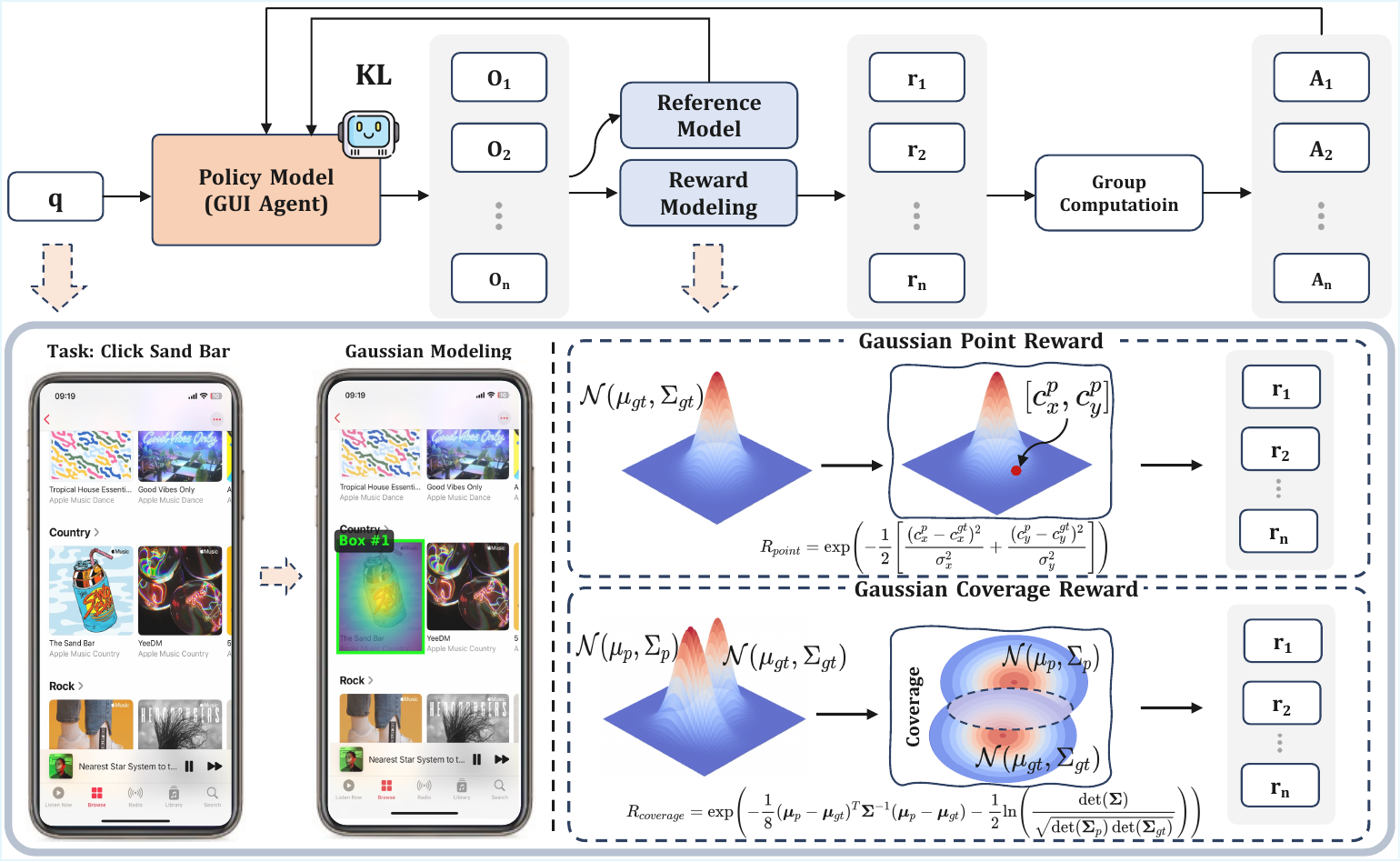}
    \caption{\textbf{GUI Gaussian Grounding Rewards (\methodname).} Our framework transforms GUI grounding through continuous Gaussian modeling. Given a task instruction and screenshot, the policy model generates multiple predictions that are evaluated using our dual reward mechanism. Gaussian Point Rewards assess localization precision while Gaussian Coverage Rewards measure spatial overlap, together providing dense learning signals that guide policy optimization.}
    \label{fig:framework}
\end{figure}

We introduce \methodname (GUI Gaussian Grounding Rewards), a principled framework that reformulates GUI grounding rewards from discrete binary signals to continuous Gaussian distributions. As illustrated in Figure~\ref{fig:framework}, our approach comprises three key innovations: (1) Gaussian point rewards that model localization precision, (2) Gaussian coverage rewards that capture spatial overlap, and (3) an adaptive variance mechanism that scales with element dimensions. This continuous formulation addresses the fundamental limitation of binary rewards by providing learning signals for near-misses through smooth gradients throughout the spatial domain.

\subsection{Problem Formulation}

GUI grounding maps natural language instructions to pixel-level targets on graphical interfaces. Given a screenshot $s$ and instruction $i$, the model must predict a bounding box $\mathbf{b}^p = [x_1^p, y_1^p, x_2^p, y_2^p]$ that localizes the element described by $i$, where $(x_1, y_1)$ and $(x_2, y_2)$ denote the top-left and bottom-right corners respectively. The ground truth is annotated as $\mathbf{b}^{gt} = [x_1^{gt}, y_1^{gt}, x_2^{gt}, y_2^{gt}]$.

In the reinforcement learning formulation, the model generates a sequence of tokens representing the predicted bounding box coordinates. The standard evaluation criterion checks whether the predicted center $(c_x^p, c_y^p) = (\frac{x_1^p + x_2^p}{2}, \frac{y_1^p + y_2^p}{2})$ falls within $\mathbf{b}^{gt}$. Our reward function $R(\mathbf{b}^p, \mathbf{b}^{gt})$ transforms this discrete success metric into continuous spatial feedback. Unlike binary rewards that provide no gradient for near-misses, \methodname generates dense learning signals that vary smoothly with prediction quality, enabling more efficient policy optimization through richer supervision.

\subsection{Gaussian Reward Modeling}

We model GUI elements as 2D Gaussian distributions to capture the continuous nature of spatial interactions. This approach transforms discrete bounding boxes into smooth probability distributions that naturally encode spatial uncertainty and provide rich gradient information.

\paragraph{Gaussian Representation.} For each GUI element with bounding box $\mathbf{b} = [x_1, y_1, x_2, y_2]$, we construct a 2D Gaussian distribution:
\begin{equation}
\mathcal{N}(\mathbf{x}; \boldsymbol{\mu}, \boldsymbol{\Sigma}) = \frac{1}{2\pi\sqrt{|\boldsymbol{\Sigma}|}} \exp\left(-\frac{1}{2}(\mathbf{x} - \boldsymbol{\mu})^T \boldsymbol{\Sigma}^{-1} (\mathbf{x} - \boldsymbol{\mu})\right)
\end{equation}
where $\mathbf{x} = (x, y)$ represents a position in the 2D interface space, $\boldsymbol{\mu} = (c_x, c_y) = (\frac{x_1 + x_2}{2}, \frac{y_1 + y_2}{2})$ is the element's geometric center, and $\boldsymbol{\Sigma} = \begin{pmatrix} \sigma_x^2 & 0 \\ 0 & \sigma_y^2 \end{pmatrix}$ is a diagonal covariance matrix. The diagonal structure assumes independence between x and y dimensions, simplifying computation while maintaining expressiveness.

\paragraph{Gaussian Point Rewards.} The point reward evaluates localization precision by measuring how well the predicted center aligns with the target element's Gaussian distribution. Given a predicted bounding box with center $\boldsymbol{\mu}_p = (c_x^p, c_y^p)$ and ground truth center $\boldsymbol{\mu}_{gt} = (c_x^{gt}, c_y^{gt})$, we compute:
\begin{equation}
R_{point} = \mathcal{N}(\boldsymbol{\mu}_p; \boldsymbol{\mu}_{gt}, \boldsymbol{\Sigma}_{gt}) = \exp\left(-\frac{1}{2}\left[\frac{(c_x^p - c_x^{gt})^2}{{\sigma_x^{gt}}^2} + \frac{(c_y^p - c_y^{gt})^2}{{\sigma_y^{gt}}^2}\right]\right)
\end{equation}
This formulation provides several key properties. First, the reward reaches its maximum value of 1 when the predicted center perfectly aligns with the ground truth. Second, it decreases smoothly and exponentially with distance, ensuring continuous gradients throughout the spatial domain. Third, the rate of decay is controlled by the variance parameters, allowing flexible adaptation to different element characteristics.

\paragraph{Gaussian Coverage Rewards.} While point rewards optimize for center alignment, GUI interactions often succeed when clicking anywhere within element boundaries. Coverage rewards capture this regional aspect by measuring the spatial overlap between predicted and target Gaussian distributions. We quantify this overlap using the Bhattacharyya coefficient:
\begin{equation}
BC(\mathcal{N}_p, \mathcal{N}_{gt}) = \int \sqrt{\mathcal{N}(\mathbf{x}; \boldsymbol{\mu}_p, \boldsymbol{\Sigma}_p) \cdot \mathcal{N}(\mathbf{x}; \boldsymbol{\mu}_{gt}, \boldsymbol{\Sigma}_{gt})} \, d\mathbf{x}
\end{equation}
For Gaussian distributions, this integral has a closed-form solution:
\begin{equation}
R_{coverage} = \exp\left(-\frac{1}{8}(\boldsymbol{\mu}_p - \boldsymbol{\mu}_{gt})^T \boldsymbol{\Sigma}^{-1} (\boldsymbol{\mu}_p - \boldsymbol{\mu}_{gt}) - \frac{1}{2}\ln\left(\frac{\det(\boldsymbol{\Sigma})}{\sqrt{\det(\boldsymbol{\Sigma}_p)\det(\boldsymbol{\Sigma}_{gt})}}\right)\right)
\end{equation}
where $\boldsymbol{\Sigma} = \frac{\boldsymbol{\Sigma}_p + \boldsymbol{\Sigma}_{gt}}{2}$ is the average covariance. The first term penalizes center misalignment weighted by the combined uncertainty, while the second term measures size and shape similarity between distributions.
\paragraph{Adaptive Variance Mechanism.} GUI elements span diverse scales, from tiny icons to full-screen panels. Fixed variance parameters would either over-constrain large elements or under-constrain small ones. We introduce an adaptive mechanism that scales variance with element dimensions:
\begin{equation}
\sigma_x = \alpha \cdot (x_2 - x_1), \quad \sigma_y = \alpha \cdot (y_2 - y_1)
\end{equation}
where $\alpha$ is a scaling factor that controls the relative influence of element size on the standard deviations. The intuition behind this scaling is straightforward: larger elements naturally tolerate greater spatial uncertainty in user interactions. A small icon requires precise targeting within a few pixels, while a large button or panel can be successfully activated across a much wider region. By making the Gaussian spread proportional to element size, we ensure that the reward function respects this natural interaction pattern. The adaptive mechanism applies to both point and coverage rewards, ensuring consistent behavior across the interface hierarchy.

\subsection{Reinforcement Learning with \methodname}

To leverage the complementary strengths of precise localization and spatial coverage, we combine both reward components:
\begin{equation}
R_{total} = \nu \cdot R_{point} + \gamma \cdot R_{coverage}
\end{equation}
where $\nu$ and $\gamma$ balance the contribution of each component. The point reward drives the model toward accurate center positioning, while the coverage reward ensures appropriate spatial extent. This dual objective mirrors human interaction patterns: users aim for element centers but can successfully interact anywhere within boundaries.

We integrate \methodname into Group Relative Policy Optimization (GRPO)~\citep{shao2024deepseekmathpushinglimitsmathematical}, which estimates advantages using multiple sampled responses. For each instruction, we sample $N$ predictions and compute their rewards under \methodname. The advantage for response $i$ is:
\begin{equation}
A_i = \frac{R_{total}(\tau_i) - \text{mean}(\{R_{total}(\tau_j)\}_{j=1}^N)}{\text{std}(\{R_{total}(\tau_j)\}_{j=1}^N)}
\end{equation}
This normalization ensures stable gradients across different element types and sizes. The policy optimization objective becomes:
\begin{equation}
\mathcal{J}(\theta) = \mathbb{E}_{\tau \sim \pi_{\theta_{old}}} \left[ \sum_{t} \min\left(r_t(\theta)A_t, \text{clip}(r_t(\theta), 1-\epsilon, 1+\epsilon)A_t\right) - \beta \mathbb{D}_{KL}[\pi_\theta \| \pi_{ref}] \right]
\end{equation}
where $r_t(\theta) = \frac{\pi_\theta(a_t|s_t)}{\pi_{\theta_{old}}(a_t|s_t)}$ is the probability ratio, $\epsilon$ controls the trust region, and $\beta$ weights the KL regularization. The continuous nature of \methodname rewards fundamentally transforms the optimization landscape. While binary rewards create a discontinuous surface with sharp cliffs at bounding box edges, our Gaussian formulation produces smooth gradients everywhere in the spatial domain. This smoothness is crucial during early training: when predictions are far from targets, the exponentially decaying Gaussian signals provide clear directional guidance toward improvement. 

\section{Experiments}
\subsection{Experiment Setup}
\textbf{Implementation Details.} We implement \methodname using Qwen2.5-VL-7B-Instruct~\citep{bai2025qwen25vltechnicalreport} as the base model within the VLM-R1 framework~\citep{shen2025vlmr1stablegeneralizabler1style}. Training is conducted on 8 NVIDIA A100-80G GPUs for one epoch with the following hyperparameters: learning rate 1e-6, global batch size 8, 8 sampled responses per instruction, and KL penalty $\beta = 0.04$. For the Gaussian reward mechanism, we set $\alpha = 0.5$. We employ Flash Attention 2~\citep{dao2023flashattention2fasterattentionbetter} and use bfloat16 precision with gradient checkpointing. During inference, we use deterministic generation with temperature 0. Unless otherwise specified, we set $\nu$ and $\gamma$ to 1.0. More training details are provided in Table \ref{tab:hyperparameters}.
The training and inference prompt templates are shown in \ref{prompt_no_thinking}. 

\textbf{Training Dataset and Evaluation Benchmarks.} 
Our training data comprises approximately 100K GUI grounding instances sampled from four major datasets: Widget Captioning~\citep{cheng2024seeclickharnessingguigrounding}, UI RefExp~\citep{bai2021uibertlearninggenericmultimodal}, ShowUI-web~\citep{lin2024showuivisionlanguageactionmodelgui}, and OmniAct~\citep{kapoor2024omniactdatasetbenchmarkenabling}, covering diverse interface types across mobile, desktop, and web platforms.
We evaluate on three benchmarks: ScreenSpot~\citep{cheng2024seeclickharnessingguigrounding} and ScreenSpot-v2~\citep{wu2024osatlasfoundationactionmodel} for general GUI grounding, and ScreenSpot-Pro~\citep{li2024screenspot-pro} for high-resolution professional software interfaces.
Following standard protocol~\citep{cheng2024seeclickharnessingguigrounding,lin2024showuivisionlanguageactionmodelgui}, predictions are considered correct when the predicted center falls within the ground truth bounding box.

\begin{table*}[!t] 
    \centering
    \small
    \begin{tabular*}{\textwidth}{@{\extracolsep{\fill}}l ccccccc c}
    \toprule
    \multirow{1}{*}{\textbf{Model}} & \multicolumn{6}{c}{\textbf{ScreenSpot v1 Accuracy (\%)}} & \multirow{1}{*}{\textbf{SSv1 Avg.}} & \multirow{1}{*}{\textbf{SSv2 Avg.}} \\
    \cmidrule(lr){2-7}
    & \multicolumn{2}{c}{Mobile} & \multicolumn{2}{c}{Desktop} & \multicolumn{2}{c}{Web} & & \\
    \cmidrule(lr){2-3} \cmidrule(lr){4-5} \cmidrule(lr){6-7}
    & Text & Icon & Text & Icon & Text & Icon & & \\
    \midrule
    \rowcolor{gray!15}
    \multicolumn{9}{l}{\textit{Proprietary Models}} \\
    GPT-4o & 30.5 & 23.2 & 20.6 & 19.4 & 11.1 & 7.8 & 18.8 & 20.1 \\
    Claude Computer Use & - & - & - & - & - & - & 83.0 & - \\
    \midrule
    \rowcolor{gray!15}
    \multicolumn{9}{l}{\textit{General Open-source Models}} \\
    Qwen2-VL-7B & 61.3 & 39.3 & 52.0 & 45.0 & 33.0 & 21.8 & 42.9 & - \\
    Qwen2.5-VL-3B & - & - & - & - & - & - & 55.5 & 80.9 \\
    Qwen2.5-VL-7B & - & - & - & - & - & - & 84.7 & 88.8 \\
    \midrule
    \rowcolor{gray!15}
    \multicolumn{9}{l}{\textit{GUI-specific Models (SFT)}} \\
    CogAgent-18B & 67.0 & 24.0 & 74.2 & 20.0 & 70.4 & 28.6 & 47.4 & - \\
    SeeClick-9.6B & 78.0 & 52.0 & 72.2 & 30.0 & 55.7 & 32.5 & 53.4 & 55.1 \\
    UGround-7B & 82.8 & 60.3 & 82.5 & 63.6 & 80.4 & 70.4 & 73.3 & 76.3 \\
    OS-Atlas-7B & 93.0 & 72.9 & 91.8 & 62.9 & 90.9 & 74.3 & 82.5 & - \\
    ShowUI-2B & 92.3 & 75.5 & 76.3 & 61.1 & 81.7 & 63.6 & 75.1 & 77.3 \\
    \textsc{Focus}-2B & 90.1 & 78.2 & 80.9 & 65.0 & 81.7 & 68.5 & 77.4 & - \\
    Aguvis-7B & 95.6 & 77.7 & 93.8 & 67.1 & 88.3 & 75.2 & 84.4 & 80.5 \\
    Aguvis-72B & 94.5 & 85.2 & 95.4 & 77.9 & 91.3 & 85.9 & 89.2 & - \\
    UI-TARS-2B & 93.0 & 75.5 & 90.7 & 68.6 & 84.3 & 74.8 & 82.3 & 84.7 \\
    UI-TARS-7B & 94.5 & 85.2 & 95.9 & 85.7 & 90.0 & 83.5 & 89.5 & 91.6 \\
    UI-TARS-72B & 94.9 & 82.5 & 89.7 & 88.6 & 88.7 & 85.0 & 88.4 & 90.3 \\
    GUI-Actor-7B & 94.9 & 82.1 & 91.8 & 80.0 & 91.3 & 85.4 & 88.3 & 92.1 \\
    \textsc{Jedi}-3B & - & - & - & - & - & - & - & 88.6 \\
    \textsc{Jedi}-7B & - & - & - & - & - & - & - & 91.7 \\
    \midrule
    \rowcolor{gray!15}
    \multicolumn{9}{l}{\textit{GUI-specific Models (RL)}} \\
    UI-R1-3B & 95.6 & 84.7 & 90.2 & 59.3 & 85.2 & 73.3 & 83.3 & 85.4 \\
    UI-R1-E-3B & 97.1 & 83.0 & 95.4 & 77.9 & 91.7 & 85.0 & 89.2 & 89.5 \\
    GUI-R1-3B & - & - & 93.8 & 64.8 & 89.6 & 72.1 & - & - \\
    GUI-R1-7B & - & - & 91.8 & 73.6 & 91.3 & 75.7 & - & - \\
    InfiGUI-R1-3B & 97.1 & 81.2 & 94.3 & 77.1 & 91.7 & 77.6 & 87.5 & - \\
    GUI-G1-3B & \textbf{98.6} & 85.8 & \textbf{96.4} & 80.7 & 91.4 & 82.3 & 90.3 & - \\
    SE-GUI-7B & - & - & - & - & - & - & 88.2 & 90.3 \\
    LPO-8B & - & - & - & - & - & - & - & 90.5 \\
    \midrule
    \rowcolor{gray!15}
    \multicolumn{9}{l}{\textit{Ours}} \\
    \methodname-7B & 96.7 & \textbf{90.8} & 95.9 & \textbf{88.6} & \textbf{90.9} & \textbf{86.9} & \textbf{92.0} & \textbf{93.3} \\
    \bottomrule
    \end{tabular*}
    \caption{Performance comparison on ScreenSpot v1 and v2. \textbf{Bold} highlights the best results, ``-'' indicates missing values due to unavailable results in the original paper, unreleased model checkpoints, and inference code.}
    \label{tab:ssv1v2}
    \vspace{-2em}
\end{table*}
\begin{table}[h]
    \centering
    \footnotesize
    \begin{tabular*}{\columnwidth}
    {@{\extracolsep{\fill}}c*{7}{c}}
    \toprule
    \multirow{2}{*}{\textbf{Reward Type}} & \multicolumn{2}{c}{\textbf{Mobile}}  & \multicolumn{2}{c}{\textbf{Desktop}} & \multicolumn{2}{c}{\textbf{Web}}     & \multirow{2}{*}{\textbf{Avg}} \\ 
    \cmidrule(lr){2-3} \cmidrule(lr){4-5} \cmidrule(lr){6-7}
                                          & \textbf{Text} & \textbf{Icon/Widget} & \textbf{Text} & \textbf{Icon/Widget} & \textbf{Text} & \textbf{Icon/Widget} &                      \\ 
    \midrule
    \rowcolor{gray!15}
    \multicolumn{8}{l}{\textit{Sparse Reward}} \\
    Point             & 97.9          & 87.2                 & 88.7          & 72.1                 & 84.9          & 79.8                 & 87.4                 \\
    IoU               & 95.9          & 86.7                 & 87.1          & 69.3                 & 88.4          & 77.3                 & 85.8                 \\
    Point + IoU       & 97.2          & 86.7                 & 88.1          & 68.6                 & 88.9          & 78.8                 & 86.5                 \\ 
    \midrule
    \rowcolor{gray!15}
    \multicolumn{8}{l}{\textit{Dense Reward}} \\
    \methodname-7B         & \textbf{98.3} & \textbf{91.9}        & \textbf{95.4} & \textbf{89.3}        & \textbf{94.0} & \textbf{87.7}        & \textbf{93.3}        \\
    \bottomrule
    \end{tabular*}
    \vspace{0.4em}
    \caption{Comparison of sparse and dense reward methods on ScreenSpot-v2.}
    \label{tab:binary_reward}
    \vspace{-2em}
\end{table}
\subsection{Main Results}
We evaluate \methodname-7B against existing methods across three benchmarks: ScreenSpot, ScreenSpot-v2, and ScreenSpot-Pro. Tables~\ref{tab:ssv1v2} and~\ref{tab:screenspot_pro} show that our method achieves state-of-the-art performance among reinforcement learning approaches.

\methodname-7B reaches 92.0\% on ScreenSpot, 93.3\% on ScreenSpot-v2, and 47.5\% on ScreenSpot-Pro, consistently outperforming all RL baselines. The most significant improvement occurs on ScreenSpot-Pro, where we surpass UI-TARS-72B by 9.4\% (47.5\% vs. 38.1\%) while using 10× fewer parameters. This efficiency gain demonstrates that continuous Gaussian rewards enable smaller models to outperform much larger counterparts through more effective optimization.

Compared to other continuous reward methods, \methodname shows clear advantages. While LPO-8B and SE-GUI-7B also employ distance-based continuous rewards, they achieve only 90.5\% and 90.3\% respectively on ScreenSpot-v2, falling short of our 93.3\%. This performance gap stems from a key insight: these methods treat GUI elements as point targets with distance decay, missing the planar nature of interface interactions. Our dual Gaussian formulation explicitly models both precise localization through point rewards and spatial extent through coverage rewards, capturing the complete interaction space that distance-only methods overlook.

The consistent improvements across diverse interface types validate the generalizability of our approach. On ScreenSpot-Pro's high-resolution professional software, we achieve 64.7\% on text elements compared to UI-TARS-72B's 50.9\%, indicating that Gaussian rewards particularly benefit tasks requiring fine spatial precision. These comprehensive improvements establish continuous Gaussian modeling as a principled foundation for GUI grounding, transforming sparse binary optimization into dense spatial learning that aligns with natural interaction patterns.
\begin{figure}[t]
    \centering
    \begin{subfigure}[b]{0.58\textwidth}
        \centering
        \includegraphics[width=\textwidth]{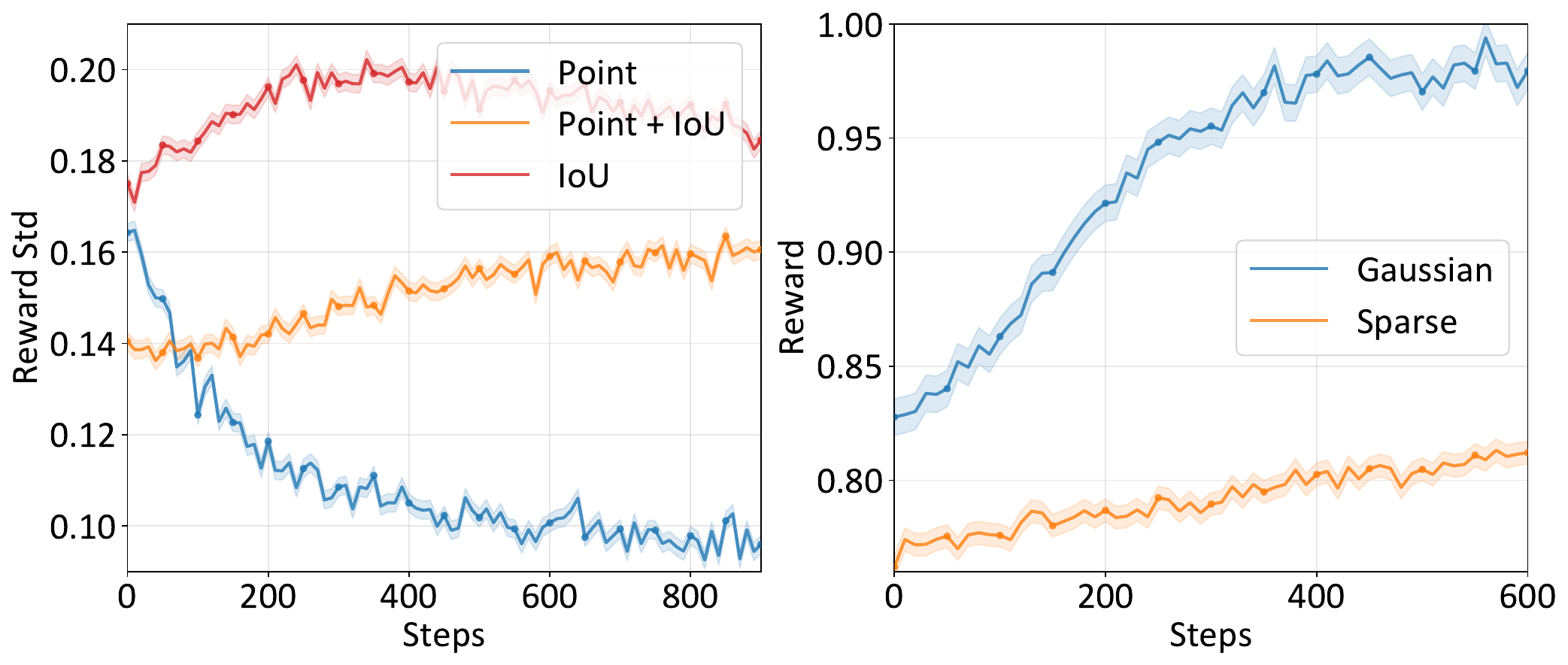}
        \caption{Sparse reward training dynamics.}
        \label{fig:sparse_ablation}
    \end{subfigure}
    \begin{subfigure}[b]{0.37\textwidth}
        \centering
        \includegraphics[width=\textwidth]{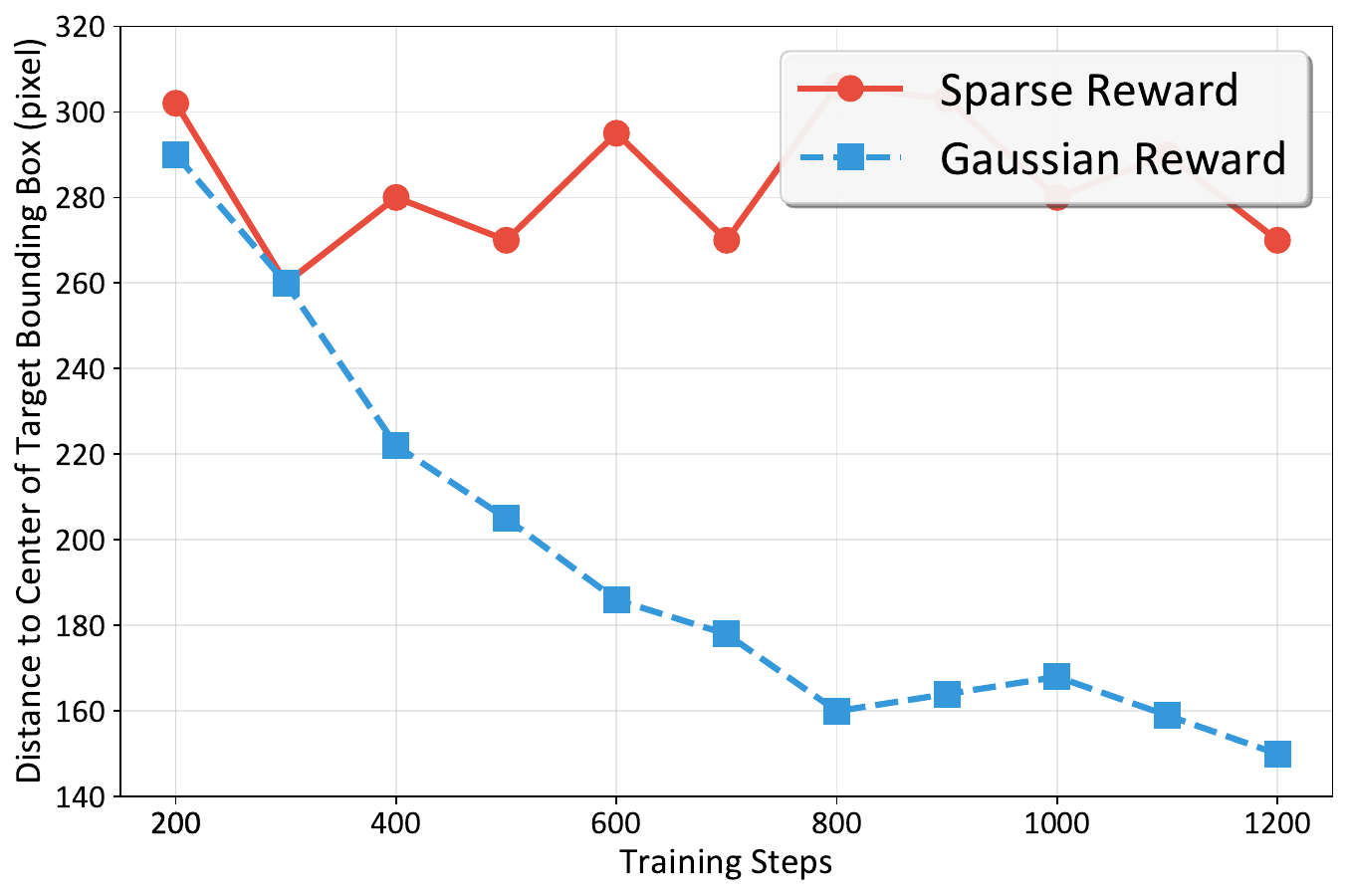}
        \caption{Distance convergence comparison.}
        \label{fig:gaussian_sparse}
    \end{subfigure}
    \caption{Reward comparison analysis. \textbf{Left:} Training dynamics of sparse reward variants (Point, IoU, Point+IoU) showing reward standard deviation and convergence patterns. \textbf{Right:} Distance to target center over training steps, where \textcolor{ourreward}{\textbf{Gaussian}} rewards demonstrate monotonic convergence while \textcolor{sparsereward}{\textbf{Sparse}} rewards exhibit erratic fluctuations.}
    \label{fig:reward_analysis}
    \vspace{-1em}
\end{figure}
\begin{table*}[!h]
    \centering
    \footnotesize
    \setlength{\tabcolsep}{0.2pt}
    \begin{tabular*}{\textwidth}{@{\extracolsep{\fill}}l *{15}{c}}
    \toprule
    \multirow{3}{*}{\textbf{Model}} & \multicolumn{2}{c}{\textbf{CAD}} & \multicolumn{2}{c}{\textbf{Dev}} & \multicolumn{2}{c}{\textbf{Creative}} & \multicolumn{2}{c}{\textbf{Scientific}} & \multicolumn{2}{c}{\textbf{Office}} & \multicolumn{2}{c}{\textbf{OS}} & \multicolumn{3}{c}{\textbf{Avg.}} \\
    \cmidrule(lr){2-3} \cmidrule(lr){4-5} \cmidrule(lr){6-7} \cmidrule(lr){8-9} \cmidrule(lr){10-11} \cmidrule(lr){12-13} \cmidrule(lr){14-16}
    & Text & Icon & Text & Icon & Text & Icon & Text & Icon & Text & Icon & Text & Icon & Text & Icon & \textbf{Avg.} \\
    \midrule
    \rowcolor{gray!15}
    \multicolumn{16}{l}{\textit{Proprietary Models}} \\
    GPT-4o & 2.0 & 0.0 & 1.3 & 0.0 & 1.0 & 0.0 & 2.1 & 0.0 & 1.1 & 0.0 & 0.0 & 0.0 & 1.3 & 0.0 & 0.8 \\
    Claude Computer Use & 14.5 & 3.7 & 22.0 & 3.9 & 25.9 & 3.4 & 33.9 & 15.8 & 30.1 & 16.3 & 11.0 & 4.5 & 23.4 & 7.1 & 17.1 \\
    \midrule
    \rowcolor{gray!15}
    \multicolumn{16}{l}{\textit{General Open-source Models}} \\
    Qwen2.5-VL-3B & 9.1 & 7.3 & 22.1 & 1.4 & 26.8 & 2.1 & 38.2 & 7.3 & 33.9 & 15.1 & 10.3 & 1.1 & 23.6 & 3.8 & 16.1 \\
    Qwen2.5-VL-7B & 16.8 & 1.6 & 46.8 & 4.1 & 35.9 & 7.7 & 49.3 & 7.3 & 52.5 & 20.8 & 37.4 & 6.7 & 38.9 & 7.1 & 26.8 \\
    \midrule
    \rowcolor{gray!15}
    \multicolumn{16}{l}{\textit{GUI-specific Models (SFT)}} \\
    SeeClick-9.6B & 2.5 & 0.0 & 0.6 & 0.0 & 1.0 & 0.0 & 3.5 & 0.0 & 1.1 & 0.0 & 2.8 & 0.0 & 1.8 & 0.0 & 1.1 \\
    \textsc{Focus}-2B & 7.6 & 3.1 & 22.8 & 1.7 & 23.7 & 1.7 & 25.0 & 7.1 & 23.2 & 7.7 & 17.8 & 2.5 & 19.8 & 3.9 & 13.3 \\
    CogAgent-18B & 7.1 & 3.1 & 14.9 & 0.7 & 9.6 & 0.0 & 22.2 & 1.8 & 13.0 & 0.0 & 5.6 & 0.0 & 12.0 & 0.8 & 7.7 \\
    Aria-UI & 7.6 & 1.6 & 16.2 & 0.0 & 23.7 & 2.1 & 27.1 & 6.4 & 20.3 & 1.9 & 4.7 & 0.0 & 17.1 & 2.0 & 11.3 \\
    OS-Atlas-7B & 12.2 & 4.7 & 33.1 & 1.4 & 28.8 & 2.8 & 37.5 & 7.3 & 33.9 & 5.7 & 27.1 & 4.5 & 28.1 & 4.0 & 18.9 \\
    ShowUI-2B & 2.5 & 0.0 & 16.9 & 1.4 & 9.1 & 0.0 & 13.2 & 7.3 & 15.3 & 7.5 & 10.3 & 2.2 & 10.8 & 2.6 & 7.7 \\
    UGround-7B & 14.2 & 1.6 & 26.6 & 2.1 & 27.3 & 2.8 & 31.9 & 2.7 & 31.6 & 11.3 & 17.8 & 0.0 & 25.0 & 2.8 & 16.5 \\
    UGround-V1-7B & 15.8 & 1.2 & 51.9 & 2.8 & 47.5 & 9.7 & 57.6 & 14.5 & 60.5 & 13.2 & 38.3 & 7.9 & 45.2 & 8.1 & 31.1 \\
    UI-TARS-2B & 17.8 & 4.7 & 47.4 & 4.1 & 42.9 & 6.3 & 56.9 & 17.3 & 50.3 & 17.0 & 21.5 & 5.6 & 39.6 & 8.4 & 27.7 \\
    UI-TARS-7B & 20.8 & 9.4 & 58.4 & 12.4 & 50.0 & 9.1 & 63.9 & 31.8 & 63.3 & 20.8 & 30.8 & 16.9 & 47.8 & 16.2 & 35.7 \\
    UI-TARS-72B& 18.8 & 12.5 & 62.9 & 17.2 & 57.1 & 15.4 & 64.6 & 20.9 & 63.3 & 26.4 & 42.1 & 15.7 & 50.9 & 17.6 & 38.1 \\
    \textsc{Jedi}-3B & 27.4 & 9.4 & 61.0 & 13.8 & 53.5 & 8.4 & 54.2 & 18.2 & 64.4 & 32.1 & 38.3 & 9.0 & 49.8 & 13.7 & 36.1 \\
    \textsc{Jedi}-7B & 38.0 & 14.1 & 42.9 & 11.0 & 50.0 & 11.9 & 72.9 & 25.5 & 75.1 & 47.2 & 33.6 & 16.9 & 52.6 & 18.2 & 39.5 \\
    GUI-Actor-7B & - & - & - & - & - & - & - & - & - & - & - & - & - & - & 44.6 \\
    \midrule
    \rowcolor{gray!15}
    \multicolumn{16}{l}{\textit{GUI-specific Models (RL)}} \\
    UI-R1-3B & 11.2 & 6.3 & 22.7 & 4.1 & 27.3 & 3.5 & 42.4 & 11.8 & 32.2 & 11.3 & 13.1 & 4.5 & 24.9 & 6.4 & 17.8 \\
    UI-R1-E-3B & 37.1 & 12.5 & 46.1 & 6.9 & 41.9 & 4.2 & 56.9 & 21.8 & 65.0 & 26.4 & 32.7 & 10.1 & - & - & 33.5 \\
    GUI-R1-3B & {26.4} & 7.8 & 33.8 & {4.8} & 40.9 & 5.6 & {61.8} & 17.3 & 53.6 & 17.0 & 28.1 & 5.6 & - & - & - \\
    GUI-R1-7B & 23.9 & 6.3 & 49.4 & {4.8} & 38.9 & {8.4} & 55.6 & 11.8 & 58.7 & {26.4} & {42.1} & {16.9} & - & - & - \\
    InfiGUI-R1-3B & 33.0 & 14.1 & 51.3 & 12.4 & 44.9 & 7.0 & 58.3 & 20.0 & 65.5 & 28.3 & 43.9 & 12.4 & 49.1 & 14.1 & 35.7 \\ 
    GUI-G1-3B & 39.6 & 9.4 & 50.7 & 10.3 & 36.6 & 11.9 & 61.8 & 30.0 & 67.2 & 32.1 & 23.5 & 10.6 & 49.5 & 16.8 & 37.1 \\
    SE-GUI-3B & 38.1 & 12.5 & {55.8} & {7.6} & {47.0} & 4.9 & 61.8 & {16.4} & {59.9} & {24.5} & {40.2} & {12.4} & {50.4} & {11.8} & {35.9} \\
    SE-GUI-7B & 51.3 & \textbf{42.2} & 68.2 & \textbf{19.3} & \textbf{57.6} & 9.1 & 75.0 & \textbf{28.2} & \textbf{78.5} & \textbf{43.4} & 49.5 & \textbf{25.8} & 63.5 & \textbf{21.0} & 47.3 \\
    \midrule
    \rowcolor{gray!15}
    \multicolumn{16}{l}{\textit{Ours}} \\
    \methodname-7B & \textbf{55.8} & 12.5 & \textbf{68.8} & 17.2 & 57.1 & \textbf{15.4} & \textbf{77.1} & 24.5 & 74.0 & 32.7 & \textbf{57.9} & 21.3 & \textbf{64.7} & 19.6 & \textbf{47.5} \\
    \bottomrule
    \end{tabular*} 
    \caption{Performance comparison of different models across various task categories based on Text, Icon, and Average scores on ScreenSpot-Pro. "-" indicates unreported results in original papers.}
    \label{tab:screenspot_pro}
    \vspace{-2.8em}
\end{table*}
\subsection{Reward Design Analysis}
\paragraph{Binary vs. Continuous Rewards.}
We investigate the fundamental differences between binary and continuous reward mechanisms by implementing three sparse baselines: Point rewards that activate when predicted centers fall within target boxes, IoU rewards that trigger when overlap exceeds 0.5, and their combination. To analyze convergence behavior, we select 10 challenging samples from ScreenSpot-v2 where initial predictions are incorrect, then track the average distance from predicted to ground truth centers across 8 sampled responses every 200 training steps.

Figure~\ref{fig:reward_analysis} exposes the critical limitations of sparse signals. Throughout training, binary rewards generate erratic optimization trajectories with severe fluctuations in both reward values and spatial convergence. The Point baseline achieves relative stability but plateaus early, while IoU rewards demonstrate particularly poor learning dynamics due to their restrictive activation threshold. Most strikingly, sparse methods show no consistent reduction in distance to target centers, oscillating wildly between 200-400 pixels without meaningful progress.
\begin{tcolorbox}
\textbf{Finding 1.} Sparse rewards create unstable training dynamics with IoU rewards showing particularly poor learning efficiency due to restrictive thresholds.
\end{tcolorbox}
In contrast, \methodname exhibits smooth monotonic convergence from 290px to 150px, demonstrating that continuous Gaussian signals fundamentally transform the optimization landscape. Table~\ref{tab:binary_reward} quantifies this advantage: \methodname achieves 93.3\% accuracy, surpassing the best sparse baseline (Point: 87.4\%) by 5.9\%. This substantial gap emerges because Gaussian rewards provide informative gradients at every spatial position, enabling models to learn from predictions at any distance from targets. Binary rewards create a discrete cliff at bounding box edges where gradient information vanishes, leaving models without guidance for improving near-miss predictions. Our continuous formulation eliminates these optimization barriers, creating smooth paths toward target elements from any starting position.
\begin{tcolorbox}
\textbf{Finding 2.} Continuous Gaussian rewards enable monotonic convergence and achieve +5.9\% performance improvement through dense spatial feedback signals.
\end{tcolorbox}
\begin{figure}[t]
    \centering
    \begin{minipage}[b]{0.49\textwidth}
        \centering
        \includegraphics[width=\textwidth]{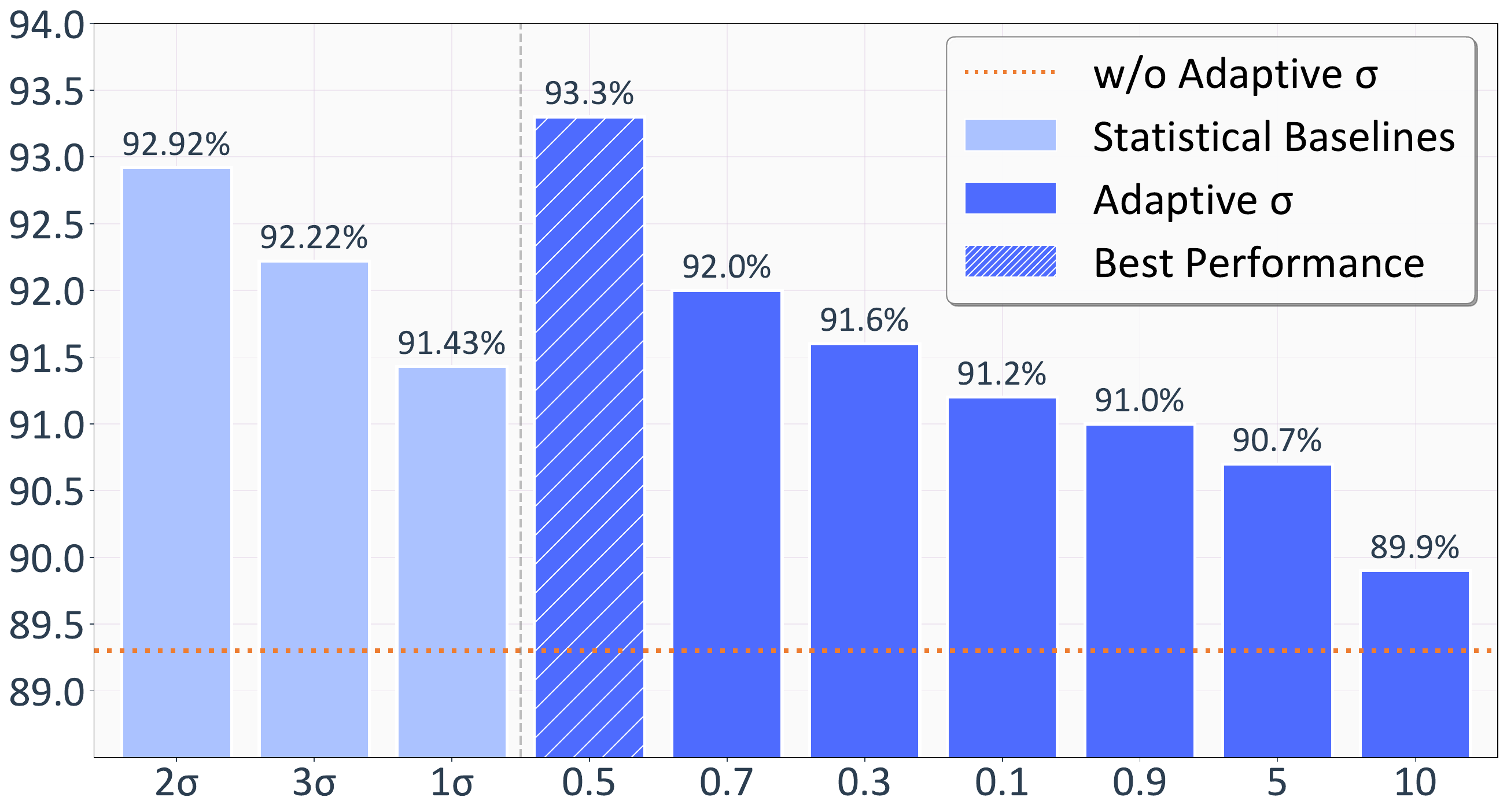}
        \caption{Hyperparameter sensitivity analysis for adaptive sigma ($\sigma$). Performance peaks at $\alpha = 0.5$ with 93.3\% accuracy on Screenspot-v2.}
        \label{fig:sigma_ablation}
    \end{minipage}
    \hfill
    \begin{minipage}[b]{0.48\textwidth}
        \centering
        \includegraphics[width=\textwidth]{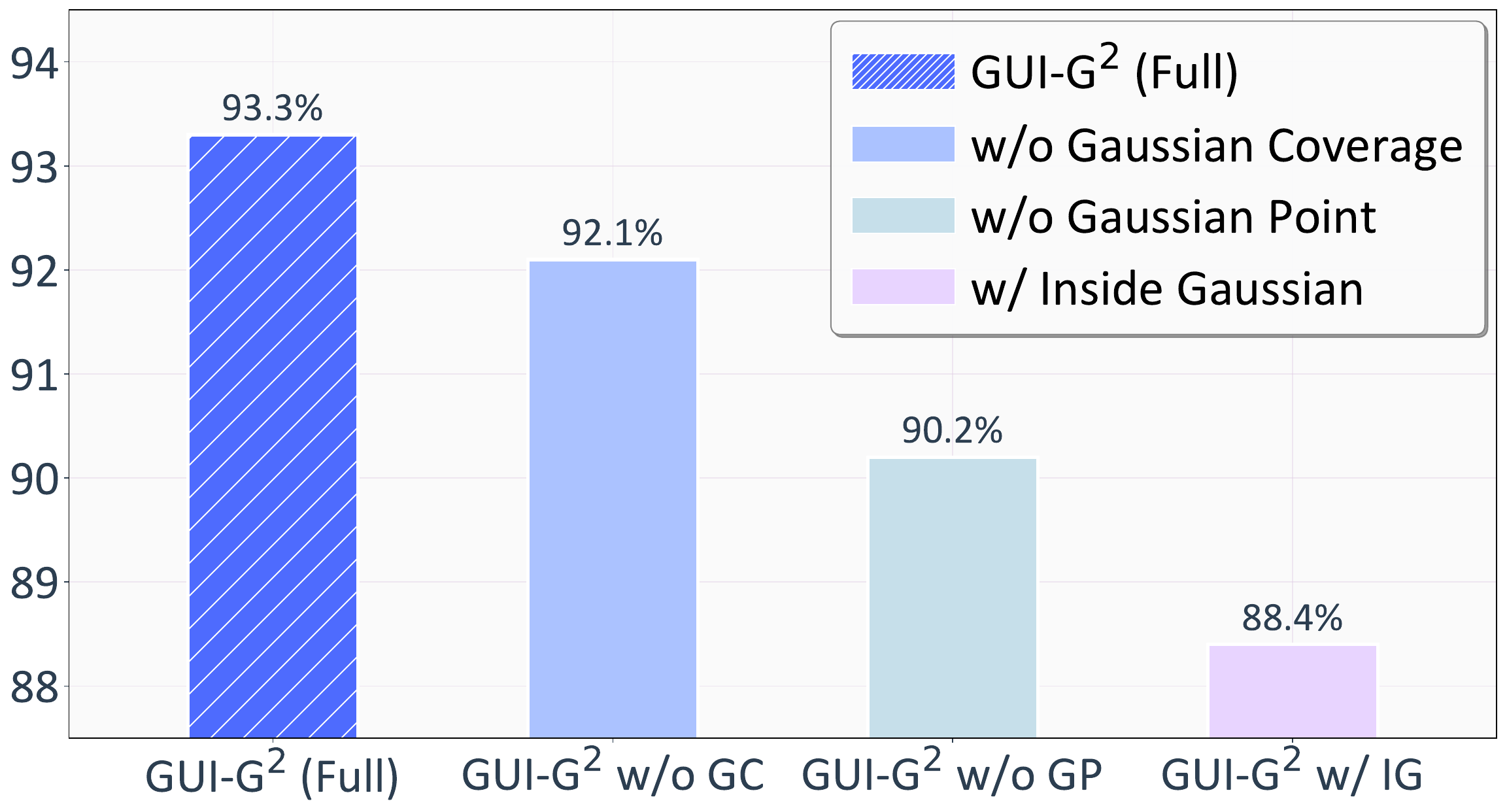}
        \caption{Ablation of Gaussian Component. Both Point and Coverage components contribute to the final 93.3\% performance.}
        \label{fig:gaussian_components}
    \end{minipage}
\end{figure}
\paragraph{Inside vs. Outside Boundary Rewards: Why Continuous Everywhere Matters.}
A natural question arises: should rewards be provided only within target boundaries or everywhere in the interface? We implement an Inside Gaussian (IG) baseline that applies our Gaussian formulation only when predictions fall within ground truth boxes, reverting to zero otherwise. Figure~\ref{fig:gaussian_components} shows \methodname outperforms IG by 4.9\% (93.3\% vs. 88.4\%), despite using identical Gaussian formulations. This reveals that restricting rewards to successful predictions, even with continuous formulations, recreates the fundamental problem of sparse signals. By providing Gaussian feedback throughout the entire interface plane, \methodname enables models to learn from every prediction, creating smooth optimization paths from any starting position to target elements.
\begin{tcolorbox}
\textbf{Finding 3.} Providing continuous Gaussian rewards everywhere in the spatial domain, rather than only within target boundaries, improves performance by 4.9\% through eliminating optimization discontinuities.
\end{tcolorbox}
\paragraph{Dual Gaussian Components: Point Precision and Spatial Coverage.}
\methodname's dual formulation addresses complementary aspects of interface interaction. Ablation studies in Figure~\ref{fig:gaussian_components} demonstrate that removing either component significantly degrades performance: 92.1\% without coverage rewards and 90.2\% without point rewards, compared to 93.3\% with both. Point rewards alone optimize for precise center localization but ignore that users can successfully click anywhere within element boundaries. Coverage rewards alone measure spatial overlap but lack the precision to guide models toward optimal clicking positions. The 1.2\% improvement from their combination confirms that effective GUI grounding requires modeling both aspects simultaneously, reflecting how humans naturally aim for element centers while accepting clicks anywhere within boundaries.
\begin{table}[t]
    \centering
    \footnotesize
    \setlength{\tabcolsep}{0pt}
    \begin{tabular*}{\columnwidth}{@{\extracolsep{\fill}}l *{8}{c}}
    \toprule
    \multirow{2}{*}{\textbf{Model}} & \multicolumn{4}{c}{\textbf{ScreenSpot Accuracy (\%)}} & \multicolumn{4}{c}{\textbf{ScreenSpot-v2 Accuracy (\%)}}      \\
    \cmidrule(lr){2-5} \cmidrule(lr){6-9}
    & Mobile & Desktop & Web & \textbf{Avg.} & Mobile & Desktop & Web & \textbf{Avg.}\\
    \midrule
    \textbf{SE-GUI-7B} & 85.6 & 91.4 & 86.5 & 88.2 & 95.2 & 87.1 & 87.0 & 90.3\\
    \textbf{\methodname-7B} & \textbf{94.0} & \textbf{92.8} & \textbf{89.0} & \textbf{92.0} & \textbf{95.6} & \textbf{92.8} & \textbf{91.1} & \textbf{93.3} \\
    \bottomrule
    \end{tabular*}
    \vspace{0.4em}
    \caption{\textbf{Gaussian vs. Distance-based Dense Rewards}. Our \methodname outperforms SE-GUI-7B on both ScreenSpot and ScreenSpot-v2 datasets, demonstrating the effectiveness of Gaussian-based dense rewards over distance-based dense reward mechanisms.}
    \label{tab:dense_reward}
    \vspace{-2em}
\end{table}
\paragraph{\methodname vs. Distance-Based Rewards.}
SE-GUI-7B represents an alternative continuous approach using normalized Euclidean distance. Table~\ref{tab:dense_reward} shows \methodname consistently outperforms SE-GUI by 3.8\% on ScreenSpot and 3.0\% on ScreenSpot-v2. This gap highlights two fundamental differences. First, SE-GUI treats elements as point targets, computing distances to centers without considering spatial extent. Second, it applies different formulas inside versus outside bounding boxes, creating gradient discontinuities at boundaries. Our unified Gaussian formulation provides smooth gradients everywhere while explicitly modeling both localization precision and spatial coverage, better capturing the continuous nature of GUI interactions.
\subsection{Ablation Studies}
\paragraph{Adaptive Variance Mechanism.} GUI elements span diverse scales from tiny icons to full-screen panels, and human clicking tolerance naturally varies with element dimensions: larger elements accommodate greater spatial uncertainty, while small icons require precise targeting. To handle this diversity, we propose an adaptive variance mechanism that scales reward distributions based on element size and validate it against multiple baselines. We implement the following configurations: (i) 1$\sigma$ Principle: $\sigma_x = \text{width}/2$, $\sigma_y = \text{height}/2$; (ii) 2$\sigma$ Principle: $\sigma_x = \text{width} \times 2$, $\sigma_y = \text{height} \times 2$; (iii) 3$\sigma$ Principle: $\sigma_x = \text{width} \times 3$, $\sigma_y = \text{height} \times 3$; and (iv) w/o adaptive $\sigma$: using fixed variance for all elements. As shown in Table~\ref{fig:sigma_ablation}, our adaptive mechanism with $\alpha=0.5$ achieves peak performance at 93.3\%, substantially outperforming fixed variance approaches (87.8\%) by +5.5 percentage points. Remarkably, this optimal value aligns with the 2$\sigma$ statistical principle (92.92\%), demonstrating that effective GUI grounding emerges from balanced spatial tolerance that neither over-constrains nor under-constrains interaction boundaries. The 1$\sigma$ principle (91.43\%) proves overly restrictive by failing to capture natural clicking variability, while the 3$\sigma$ principle (92.22\%) shows that excessive tolerance dilutes localization precision. Our adaptive mechanism's superiority over even the optimal 2$\sigma$ baseline reveals that personalized calibration based on individual element characteristics provides the optimal balance between spatial flexibility and targeting precision.
\begin{table}[t]
\centering
\begin{minipage}{0.48\textwidth}
    \centering
    \small
    \begin{tabular}{lccc}
    \toprule
    \textbf{Configuration} & $\nu$ & $\gamma$ & \textbf{Acc (\%)} \\
    \midrule
    \methodname & 1.0 & 1.0 & \textbf{93.3} \\
    \methodname[Format] & 1.0 & 1.0 & 93.2 \\
    \methodname[GP] & 0.8 & 0.2 & 92.2 \\
    \methodname[GC] & 0.2 & 0.8 & 91.8 \\
    \bottomrule
    \end{tabular}
    \vspace{0.4em}
    \caption{Reward weighting configurations.}
    \label{tab:reward_weighting}
\end{minipage}
\hfill
\begin{minipage}{0.48\textwidth}
    \centering
    \small
    \begin{tabular}{lcc}
    \toprule
    \textbf{Configuration} & \textbf{Accuracy (\%)} & \textbf{Tokens} \\
    \midrule
    Thinking & 88.7 & 130 \\
    \rowcolor{gray!15}
    No Thinking & \textbf{93.3} & \textbf{16} \\
    \midrule
    $\Delta$ & \textcolor{citeblue}{+4.6} & \textcolor{citeblue}{-114} \\
    Relative & \textcolor{citeblue}{+5.2\%} & \textcolor{citeblue}{-87.7\%} \\
    \bottomrule
    \end{tabular}
    \vspace{0.4em}
    \caption{Thinking vs. No Thinking Analysis.}
    \label{tab:thinking_detailed}
\end{minipage}
\vspace{-2.6em}
\end{table}
\paragraph{Balancing Point and Coverage Rewards.} To evaluate the impact of different weighting schemes between Gaussian point and Gaussian coverage rewards,
we perform ablation experiments with the following configurations:
(i) \methodname: our original model 
with $R = 1.0 \times R_{point} + 1.0 \times R_{coverage}$; 
(ii) \methodname[GP]: Point-dominant weighting with $R = 0.8 \times R_{point} + 0.2 \times R_{coverage}$; 
(iii) \methodname[GC]: Coverage-dominant weighting with $R = 0.2 \times R_{point} + 0.8 \times R_{coverage}$; 
(iv) \methodname[Format]: our original model with additional format reward that assigns reward 1 when the model outputs exactly four numerical coordinates in the required format $[x1,y1,x2,y2]$ and 0 otherwise. As shown in Table~\ref{tab:reward_weighting}, equal weighting (1.0 each) achieves optimal performance at 93.3\%, outperforming both point-dominant (92.2\%) and coverage-dominant (91.8\%) configurations. This demonstrates that effective GUI grounding requires simultaneous optimization of precise localization and spatial overlap modeling with balanced importance. 
\begin{tcolorbox}
\textbf{Finding 4.}
Balanced weighting of point and coverage rewards (1.0 each) achieves optimal performance, while format rewards provide minimal benefit.
\end{tcolorbox}
\paragraph{Thinking vs. No Thinking Grounding.} 
Most previous methods adopt the R1-style reasoning paradigm directly~\citep{luo2025guir1generalistr1style,liu2025infiguir1advancingmultimodalgui}, following the success of reasoning-based models in other domains. However, this widespread adoption raises a fundamental question: is explicit reasoning truly beneficial for GUI grounding tasks? To investigate whether GUI grounding is suitable for thinking-based optimization, we conduct controlled experiments comparing thinking versus non-thinking approaches. Both configurations utilize \methodname reward mechanism, with the thinking model additionally receiving format rewards for proper usage of thinking tag. 
Training prompts detailed in Appendix~\ref{sec:training_details}. As shown Table~\ref{tab:thinking_detailed},
our experiments demonstrate that \textbf{explicit reasoning significantly impairs GUI grounding performance.} The non-thinking approach achieves 93.3\% on ScreenSpot-v2, substantially outperforming the thinking approach at 88.7\%—a +5.3\% improvement while using 76.9\% fewer tokens.
This counterintuitive finding suggests that GUI grounding is fundamentally a perceptual task relying on immediate visual pattern recognition rather than step-by-step analysis. The performance degradation likely occurs because reasoning tokens compete with visual representations for attention, interfering with crucial visual features essential for accurate element localization.
\begin{tcolorbox}
\textbf{Finding 5.}
Explicit reasoning processes significantly harm GUI grounding performance.
\end{tcolorbox}

\section{Conclusion}
In this work, we propose \methodname, a principled reward modeling framework that reconceptualizes GUI grounding as a continuous spatial optimization task. Unlike traditional reinforcement learning approaches that rely on sparse binary rewards, \methodname leverages Gaussian point rewards and Gaussian coverage rewards to provide dense, geometrically-aware feedback signals. By modeling GUI elements as 2D Gaussian distributions and introducing an adaptive variance mechanism, our method captures both fine-grained localization precision and spatial coverage characteristics, which enables more efficient learning and better generalization. Evaluated on three benchmarks—ScreenSpot, ScreenSpot-v2, and ScreenSpot-Pro. \methodname-7B outperforms state-of-the-art models, achieving up to 24.7\% improvement over UI-TARS-72B on high-resolution professional interfaces. These results establish \methodname as a robust and effective solution for spatial reasoning in GUI interaction tasks.

\bibliographystyle{iclr2026_conference}
\bibliography{ref}

\begin{thebibliography}{51}
\providecommand{\natexlab}[1]{#1}
\providecommand{\url}[1]{\texttt{#1}}
\expandafter\ifx\csname urlstyle\endcsname\relax
  \providecommand{\doi}[1]{doi: #1}\else
  \providecommand{\doi}{doi: \begingroup \urlstyle{rm}\Url}\fi

\bibitem[and(1992)]{mackenzie1992fitts}
I.~Scott~MacKenzie and.
\newblock Fitts' law as a research and design tool in human-computer interaction.
\newblock \emph{Human–Computer Interaction}, 7\penalty0 (1):\penalty0 91--139, 1992.
\newblock \doi{10.1207/s15327051hci0701\_3}.
\newblock URL \url{https://doi.org/10.1207/s15327051hci0701_3}.

\bibitem[Anthropic(2024)]{anthropic2024cuda}
Anthropic.
\newblock Claude computer use.
\newblock Available at: https://www.anthropic.com/news/developing-computer-use, 2024.

\bibitem[Bai et~al.(2021)Bai, Zang, Xu, Sunkara, Rastogi, Chen, and y~Arcas]{bai2021uibertlearninggenericmultimodal}
Chongyang Bai, Xiaoxue Zang, Ying Xu, Srinivas Sunkara, Abhinav Rastogi, Jindong Chen, and Blaise~Aguera y~Arcas.
\newblock Uibert: Learning generic multimodal representations for ui understanding, 2021.
\newblock URL \url{https://arxiv.org/abs/2107.13731}.

\bibitem[Bai et~al.(2025)Bai, Chen, Liu, Wang, Ge, Song, Dang, Wang, Wang, Tang, Zhong, Zhu, Yang, Li, Wan, Wang, Ding, Fu, Xu, Ye, Zhang, Xie, Cheng, Zhang, Yang, Xu, and Lin]{bai2025qwen25vltechnicalreport}
Shuai Bai, Keqin Chen, Xuejing Liu, Jialin Wang, Wenbin Ge, Sibo Song, Kai Dang, Peng Wang, Shijie Wang, Jun Tang, Humen Zhong, Yuanzhi Zhu, Mingkun Yang, Zhaohai Li, Jianqiang Wan, Pengfei Wang, Wei Ding, Zheren Fu, Yiheng Xu, Jiabo Ye, Xi~Zhang, Tianbao Xie, Zesen Cheng, Hang Zhang, Zhibo Yang, Haiyang Xu, and Junyang Lin.
\newblock Qwen2.5-vl technical report, 2025.
\newblock URL \url{https://arxiv.org/abs/2502.13923}.

\bibitem[Cheng et~al.(2024)Cheng, Sun, Chu, Xu, Li, Zhang, and Wu]{cheng2024seeclickharnessingguigrounding}
Kanzhi Cheng, Qiushi Sun, Yougang Chu, Fangzhi Xu, Yantao Li, Jianbing Zhang, and Zhiyong Wu.
\newblock Seeclick: Harnessing gui grounding for advanced visual gui agents, 2024.
\newblock URL \url{https://arxiv.org/abs/2401.10935}.

\bibitem[Chu et~al.(2025)Chu, Zhai, Yang, Tong, Xie, Schuurmans, Le, Levine, and Ma]{chu2025sftmemorizesrlgeneralizes}
Tianzhe Chu, Yuexiang Zhai, Jihan Yang, Shengbang Tong, Saining Xie, Dale Schuurmans, Quoc~V. Le, Sergey Levine, and Yi~Ma.
\newblock Sft memorizes, rl generalizes: A comparative study of foundation model post-training, 2025.
\newblock URL \url{https://arxiv.org/abs/2501.17161}.

\bibitem[Dao(2023)]{dao2023flashattention2fasterattentionbetter}
Tri Dao.
\newblock Flashattention-2: Faster attention with better parallelism and work partitioning, 2023.
\newblock URL \url{https://arxiv.org/abs/2307.08691}.

\bibitem[DeepSeek-AI(2025)]{deepseekai2025deepseekr1incentivizingreasoningcapability}
DeepSeek-AI.
\newblock Deepseek-r1: Incentivizing reasoning capability in llms via reinforcement learning, 2025.
\newblock URL \url{https://arxiv.org/abs/2501.12948}.

\bibitem[Du et~al.(2020)Du, Li, Guo, Yin, Liu, Zhou, Bai, Yu, Yang, Dang, and Wang]{du2020ppocrpracticalultralightweight}
Yuning Du, Chenxia Li, Ruoyu Guo, Xiaoting Yin, Weiwei Liu, Jun Zhou, Yifan Bai, Zilin Yu, Yehua Yang, Qingqing Dang, and Haoshuang Wang.
\newblock Pp-ocr: A practical ultra lightweight ocr system, 2020.
\newblock URL \url{https://arxiv.org/abs/2009.09941}.

\bibitem[Feng et~al.(2025)Feng, Gong, Li, Guo, Wang, Peng, Wu, Zhang, Wang, and Yue]{feng2025videor1reinforcingvideoreasoning}
Kaituo Feng, Kaixiong Gong, Bohao Li, Zonghao Guo, Yibing Wang, Tianshuo Peng, Junfei Wu, Xiaoying Zhang, Benyou Wang, and Xiangyu Yue.
\newblock Video-r1: Reinforcing video reasoning in mllms, 2025.
\newblock URL \url{https://arxiv.org/abs/2503.21776}.

\bibitem[Fitts(1954)]{fitts1954information}
P.~M. Fitts.
\newblock The information capacity of the human motor system in controlling the amplitude of movement.
\newblock \emph{Journal of Experimental PSychology}, 74:\penalty0 381--391, 1954.

\bibitem[Gou et~al.(2024)Gou, Wang, Zheng, Xie, Chang, Shu, Sun, and Su]{gou2024navigatingdigitalworldhumans}
Boyu Gou, Ruohan Wang, Boyuan Zheng, Yanan Xie, Cheng Chang, Yiheng Shu, Huan Sun, and Yu~Su.
\newblock Navigating the digital world as humans do: Universal visual grounding for gui agents, 2024.
\newblock URL \url{https://arxiv.org/abs/2410.05243}.

\bibitem[Hong et~al.(2024)Hong, Wang, Lv, Xu, Yu, Ji, Wang, Wang, Zhang, Li, Xu, Dong, Ding, and Tang]{hong2024cogagentvisuallanguagemodel}
Wenyi Hong, Weihan Wang, Qingsong Lv, Jiazheng Xu, Wenmeng Yu, Junhui Ji, Yan Wang, Zihan Wang, Yuxuan Zhang, Juanzi Li, Bin Xu, Yuxiao Dong, Ming Ding, and Jie Tang.
\newblock Cogagent: A visual language model for gui agents, 2024.
\newblock URL \url{https://arxiv.org/abs/2312.08914}.

\bibitem[Jiang et~al.(2025)Jiang, Zhuang, Song, Yang, Zhou, and Zhang]{jiang2025appagentxevolvingguiagents}
Wenjia Jiang, Yangyang Zhuang, Chenxi Song, Xu~Yang, Joey~Tianyi Zhou, and Chi Zhang.
\newblock Appagentx: Evolving gui agents as proficient smartphone users.
\newblock 2025.
\newblock URL \url{https://arxiv.org/abs/2503.02268}.

\bibitem[Kapoor et~al.(2024)Kapoor, Butala, Russak, Koh, Kamble, Alshikh, and Salakhutdinov]{kapoor2024omniactdatasetbenchmarkenabling}
Raghav Kapoor, Yash~Parag Butala, Melisa Russak, Jing~Yu Koh, Kiran Kamble, Waseem Alshikh, and Ruslan Salakhutdinov.
\newblock Omniact: A dataset and benchmark for enabling multimodal generalist autonomous agents for desktop and web.
\newblock 2024.
\newblock URL \url{https://arxiv.org/abs/2402.17553}.

\bibitem[Kirillov et~al.(2023)Kirillov, Mintun, Ravi, Mao, Rolland, Gustafson, Xiao, Whitehead, Berg, Lo, Dollár, and Girshick]{kirillov2023segment}
Alexander Kirillov, Eric Mintun, Nikhila Ravi, Hanzi Mao, Chloe Rolland, Laura Gustafson, Tete Xiao, Spencer Whitehead, Alexander~C. Berg, Wan-Yen Lo, Piotr Dollár, and Ross Girshick.
\newblock Segment anything, 2023.
\newblock URL \url{https://arxiv.org/abs/2304.02643}.

\bibitem[Li et~al.(2025)Li, Meng, Lin, Luo, Tian, Ma, Huang, and Chua]{li2024screenspot-pro}
Kaixin Li, Ziyang Meng, Hongzhan Lin, Ziyang Luo, Yuchen Tian, Jing Ma, Zhiyong Huang, and Tat-Seng Chua.
\newblock Screenspot-pro: Gui grounding for professional high-resolution computer use, 2025.

\bibitem[Li et~al.(2024)Li, Zhang, Yang, Fu, Cheng, Chen, Chen, and Wei]{li2024appagentv2advancedagent}
Yanda Li, Chi Zhang, Wanqi Yang, Bin Fu, Pei Cheng, Xin Chen, Ling Chen, and Yunchao Wei.
\newblock Appagent v2: Advanced agent for flexible mobile interactions, 2024.
\newblock URL \url{https://arxiv.org/abs/2408.11824}.

\bibitem[Lin et~al.(2024)Lin, Li, Gao, Yang, Wu, Bai, Lei, Wang, and Shou]{lin2024showuivisionlanguageactionmodelgui}
Kevin~Qinghong Lin, Linjie Li, Difei Gao, Zhengyuan Yang, Shiwei Wu, Zechen Bai, Weixian Lei, Lijuan Wang, and Mike~Zheng Shou.
\newblock Showui: One vision-language-action model for gui visual agent, 2024.
\newblock URL \url{https://arxiv.org/abs/2411.17465}.

\bibitem[Liu et~al.(2024)Liu, Qin, Liang, Dong, Lai, Zhang, Zhao, Iong, Sun, Wang, Gao, Shan, Liu, Zhang, Yao, Cheng, Yao, Zhao, Liu, Liu, Chen, Yang, Yang, Xu, Yang, Wang, Xu, Qi, Dong, and Tang]{liu2024autoglmautonomousfoundationagents}
Xiao Liu, Bo~Qin, Dongzhu Liang, Guang Dong, Hanyu Lai, Hanchen Zhang, Hanlin Zhao, Iat~Long Iong, Jiadai Sun, Jiaqi Wang, Junjie Gao, Junjun Shan, Kangning Liu, Shudan Zhang, Shuntian Yao, Siyi Cheng, Wentao Yao, Wenyi Zhao, Xinghan Liu, Xinyi Liu, Xinying Chen, Xinyue Yang, Yang Yang, Yifan Xu, Yu~Yang, Yujia Wang, Yulin Xu, Zehan Qi, Yuxiao Dong, and Jie Tang.
\newblock Autoglm: Autonomous foundation agents for guis.
\newblock 2024.
\newblock URL \url{https://arxiv.org/abs/2411.00820}.

\bibitem[Liu et~al.(2025{\natexlab{a}})Liu, Wang, Ma, and Zhang]{liu2025videocompressioncommanderplugandplay}
Xuyang Liu, Yiyu Wang, Junpeng Ma, and Linfeng Zhang.
\newblock Video compression commander: Plug-and-play inference acceleration for video large language models, 2025{\natexlab{a}}.
\newblock URL \url{https://arxiv.org/abs/2505.14454}.

\bibitem[Liu et~al.(2025{\natexlab{b}})Liu, Wang, Han, Wang, Yuan, Song, Zheng, Zhang, Huang, and Chen]{liu2025globalcompressioncommanderplugandplay}
Xuyang Liu, Ziming Wang, Yuhang Han, Yingyao Wang, Jiale Yuan, Jun Song, Bo~Zheng, Linfeng Zhang, Siteng Huang, and Honggang Chen.
\newblock Global compression commander: Plug-and-play inference acceleration for high-resolution large vision-language models, 2025{\natexlab{b}}.
\newblock URL \url{https://arxiv.org/abs/2501.05179}.

\bibitem[Liu et~al.(2025{\natexlab{c}})Liu, Wen, Wang, Chen, Tao, Wang, Jin, Zou, Wang, Liao, Zheng, Chen, Li, Hu, He, and Zhang]{liu2025shiftingaiefficiencymodelcentric}
Xuyang Liu, Zichen Wen, Shaobo Wang, Junjie Chen, Zhishan Tao, Yubo Wang, Xiangqi Jin, Chang Zou, Yiyu Wang, Chenfei Liao, Xu~Zheng, Honggang Chen, Weijia Li, Xuming Hu, Conghui He, and Linfeng Zhang.
\newblock Shifting ai efficiency from model-centric to data-centric compression, 2025{\natexlab{c}}.
\newblock URL \url{https://arxiv.org/abs/2505.19147}.

\bibitem[Liu et~al.(2025{\natexlab{d}})Liu, Li, Xie, Hu, Han, Zhang, Yang, and Wu]{liu2025infiguir1advancingmultimodalgui}
Yuhang Liu, Pengxiang Li, Congkai Xie, Xavier Hu, Xiaotian Han, Shengyu Zhang, Hongxia Yang, and Fei Wu.
\newblock Infigui-r1: Advancing multimodal gui agents from reactive actors to deliberative reasoners.
\newblock 2025{\natexlab{d}}.
\newblock URL \url{https://arxiv.org/abs/2504.14239}.

\bibitem[Lu et~al.(2024)Lu, Yang, Shen, and Awadallah]{lu2024omniparserpurevisionbased}
Yadong Lu, Jianwei Yang, Yelong Shen, and Ahmed Awadallah.
\newblock Omniparser for pure vision based gui agent, 2024.
\newblock URL \url{https://arxiv.org/abs/2408.00203}.

\bibitem[Lu et~al.(2025)Lu, Chai, Guo, Yin, Liu, Wang, Xiao, Ren, Xiong, and Li]{lu2025uir1enhancingefficientaction}
Zhengxi Lu, Yuxiang Chai, Yaxuan Guo, Xi~Yin, Liang Liu, Hao Wang, Han Xiao, Shuai Ren, Guanjing Xiong, and Hongsheng Li.
\newblock Ui-r1: Enhancing efficient action prediction of gui agents by reinforcement learning.
\newblock 2025.
\newblock URL \url{https://arxiv.org/abs/2503.21620}.

\bibitem[Luo et~al.(2025)Luo, Wang, He, and Xia]{luo2025guir1generalistr1style}
Run Luo, Lu~Wang, Wanwei He, and Xiaobo Xia.
\newblock Gui-r1 : A generalist r1-style vision-language action model for gui agents.
\newblock 2025.
\newblock URL \url{https://arxiv.org/abs/2504.10458}.

\bibitem[OpenAI(2024)]{openai2024gpt4o}
OpenAI.
\newblock Introducing gpt-4o.
\newblock Available at: https://openai.com/index/hello-gpt-4o, 2024.

\bibitem[Qin et~al.(2025)Qin, Ye, Fang, Wang, Liang, Tian, Zhang, Li, Li, Huang, Zhong, Li, Yang, Miao, Lin, Liu, Jiang, Ma, Li, Xiao, Cai, Li, Zheng, Jin, Li, Zhou, Wang, Chen, Li, Yang, Liu, Lin, Peng, Liu, and Shi]{qin2025uitarspioneeringautomatedgui}
Yujia Qin, Yining Ye, Junjie Fang, Haoming Wang, Shihao Liang, Shizuo Tian, Junda Zhang, Jiahao Li, Yunxin Li, Shijue Huang, Wanjun Zhong, Kuanye Li, Jiale Yang, Yu~Miao, Woyu Lin, Longxiang Liu, Xu~Jiang, Qianli Ma, Jingyu Li, Xiaojun Xiao, Kai Cai, Chuang Li, Yaowei Zheng, Chaolin Jin, Chen Li, Xiao Zhou, Minchao Wang, Haoli Chen, Zhaojian Li, Haihua Yang, Haifeng Liu, Feng Lin, Tao Peng, Xin Liu, and Guang Shi.
\newblock Ui-tars: Pioneering automated gui interaction with native agents, 2025.
\newblock URL \url{https://arxiv.org/abs/2501.12326}.

\bibitem[Rawles et~al.(2023)Rawles, Li, Rodriguez, Riva, and Lillicrap]{rawles2023androidwildlargescaledataset}
Christopher Rawles, Alice Li, Daniel Rodriguez, Oriana Riva, and Timothy Lillicrap.
\newblock Android in the wild: A large-scale dataset for android device control.
\newblock 2023.
\newblock URL \url{https://arxiv.org/abs/2307.10088}.

\bibitem[Shao et~al.(2025)Shao, Li, Xin, Geng, Wang, Oh, Du, Lambert, Min, Krishna, Tsvetkov, Hajishirzi, Koh, and Zettlemoyer]{shao2025spuriousrewardsrethinkingtraining}
Rulin Shao, Shuyue~Stella Li, Rui Xin, Scott Geng, Yiping Wang, Sewoong Oh, Simon~Shaolei Du, Nathan Lambert, Sewon Min, Ranjay Krishna, Yulia Tsvetkov, Hannaneh Hajishirzi, Pang~Wei Koh, and Luke Zettlemoyer.
\newblock Spurious rewards: Rethinking training signals in rlvr, 2025.
\newblock URL \url{https://arxiv.org/abs/2506.10947}.

\bibitem[Shao et~al.(2024)Shao, Wang, Zhu, Xu, Song, Bi, Zhang, Zhang, Li, Wu, and Guo]{shao2024deepseekmathpushinglimitsmathematical}
Zhihong Shao, Peiyi Wang, Qihao Zhu, Runxin Xu, Junxiao Song, Xiao Bi, Haowei Zhang, Mingchuan Zhang, Y.~K. Li, Y.~Wu, and Daya Guo.
\newblock Deepseekmath: Pushing the limits of mathematical reasoning in open language models, 2024.
\newblock URL \url{https://arxiv.org/abs/2402.03300}.

\bibitem[Shen et~al.(2025)Shen, Liu, Li, Fang, Ma, Liao, Shen, Zhang, Zhao, Zhang, Xu, and Zhao]{shen2025vlmr1stablegeneralizabler1style}
Haozhan Shen, Peng Liu, Jingcheng Li, Chunxin Fang, Yibo Ma, Jiajia Liao, Qiaoli Shen, Zilun Zhang, Kangjia Zhao, Qianqian Zhang, Ruochen Xu, and Tiancheng Zhao.
\newblock Vlm-r1: A stable and generalizable r1-style large vision-language model, 2025.
\newblock URL \url{https://arxiv.org/abs/2504.07615}.

\bibitem[Shen et~al.(2023)Shen, Song, Tan, Li, Lu, and Zhuang]{shen2023hugginggptsolvingaitasks}
Yongliang Shen, Kaitao Song, Xu~Tan, Dongsheng Li, Weiming Lu, and Yueting Zhuang.
\newblock Hugginggpt: Solving ai tasks with chatgpt and its friends in hugging face, 2023.
\newblock URL \url{https://arxiv.org/abs/2303.17580}.

\bibitem[Sun et~al.(2025)Sun, Cheng, Ding, Jin, Wang, Xu, Wu, Jia, Chen, Liu, Kao, Li, He, Qiao, and Wu]{sun2025osgenesisautomatingguiagent}
Qiushi Sun, Kanzhi Cheng, Zichen Ding, Chuanyang Jin, Yian Wang, Fangzhi Xu, Zhenyu Wu, Chengyou Jia, Liheng Chen, Zhoumianze Liu, Ben Kao, Guohao Li, Junxian He, Yu~Qiao, and Zhiyong Wu.
\newblock Os-genesis: Automating gui agent trajectory construction via reverse task synthesis, 2025.
\newblock URL \url{https://arxiv.org/abs/2412.19723}.

\bibitem[Tang et~al.(2025{\natexlab{a}})Tang, Shen, Zhang, Chen, Hou, Zhang, Zhang, Song, Lu, and Zhuang]{tang2025thinktwiceclickonce}
Fei Tang, Yongliang Shen, Hang Zhang, Siqi Chen, Guiyang Hou, Wenqi Zhang, Wenqiao Zhang, Kaitao Song, Weiming Lu, and Yueting Zhuang.
\newblock Think twice, click once: Enhancing gui grounding via fast and slow systems.
\newblock 2025{\natexlab{a}}.
\newblock URL \url{https://arxiv.org/abs/2503.06470}.

\bibitem[Tang et~al.(2025{\natexlab{b}})Tang, Xu, Zhang, Chen, Wu, Shen, Zhang, Hou, Tan, Yan, Song, Shao, Lu, Xiao, and Zhuang]{tang2025surveymllmbasedguiagents}
Fei Tang, Haolei Xu, Hang Zhang, Siqi Chen, Xingyu Wu, Yongliang Shen, Wenqi Zhang, Guiyang Hou, Zeqi Tan, Yuchen Yan, Kaitao Song, Jian Shao, Weiming Lu, Jun Xiao, and Yueting Zhuang.
\newblock A survey on (m)llm-based gui agents.
\newblock 2025{\natexlab{b}}.
\newblock URL \url{https://arxiv.org/abs/2504.13865}.

\bibitem[Tang et~al.(2025{\natexlab{c}})Tang, Xia, Wu, Hu, Chen, Chen, Xu, Wu, Lu, Ma, Lu, and Chen]{tang2025lpoaccurateguiagent}
Jiaqi Tang, Yu~Xia, Yi-Feng Wu, Yuwei Hu, Yuhui Chen, Qing-Guo Chen, Xiaogang Xu, Xiangyu Wu, Hao Lu, Yanqing Ma, Shiyin Lu, and Qifeng Chen.
\newblock Lpo: Towards accurate gui agent interaction via location preference optimization, 2025{\natexlab{c}}.
\newblock URL \url{https://arxiv.org/abs/2506.09373}.

\bibitem[Wang et~al.(2024{\natexlab{a}})Wang, Xu, Jia, Zhang, Yan, Shen, Zhang, Huang, and Sang]{wang2024mobileagentv2mobiledeviceoperation}
Junyang Wang, Haiyang Xu, Haitao Jia, Xi~Zhang, Ming Yan, Weizhou Shen, Ji~Zhang, Fei Huang, and Jitao Sang.
\newblock Mobile-agent-v2: Mobile device operation assistant with effective navigation via multi-agent collaboration, 2024{\natexlab{a}}.
\newblock URL \url{https://arxiv.org/abs/2406.01014}.

\bibitem[Wang et~al.(2024{\natexlab{b}})Wang, Xu, Ye, Yan, Shen, Zhang, Huang, and Sang]{wang2024mobileagentautonomousmultimodalmobile}
Junyang Wang, Haiyang Xu, Jiabo Ye, Ming Yan, Weizhou Shen, Ji~Zhang, Fei Huang, and Jitao Sang.
\newblock Mobile-agent: Autonomous multi-modal mobile device agent with visual perception, 2024{\natexlab{b}}.
\newblock URL \url{https://arxiv.org/abs/2401.16158}.

\bibitem[Wang et~al.(2025)Wang, Xu, Wang, Zhang, Yan, Zhang, Huang, and Ji]{wang2025mobileagenteselfevolvingmobileassistant}
Zhenhailong Wang, Haiyang Xu, Junyang Wang, Xi~Zhang, Ming Yan, Ji~Zhang, Fei Huang, and Heng Ji.
\newblock Mobile-agent-e: Self-evolving mobile assistant for complex tasks, 2025.
\newblock URL \url{https://arxiv.org/abs/2501.11733}.

\bibitem[Wu et~al.(2025)Wu, Cheng, Yang, Zhang, Yang, Jiang, Mu, Peng, Qiao, Tan, et~al.]{guiactor}
Qianhui Wu, Kanzhi Cheng, Rui Yang, Chaoyun Zhang, Jianwei Yang, Huiqiang Jiang, Jian Mu, Baolin Peng, Bo~Qiao, Reuben Tan, et~al.
\newblock Gui-actor: Coordinate-free visual grounding for gui agents.
\newblock \emph{arXiv preprint arXiv:2506.03143}, 2025.

\bibitem[Wu et~al.(2024)Wu, Wu, Xu, Wang, Sun, Jia, Cheng, Ding, Chen, Liang, and Qiao]{wu2024osatlasfoundationactionmodel}
Zhiyong Wu, Zhenyu Wu, Fangzhi Xu, Yian Wang, Qiushi Sun, Chengyou Jia, Kanzhi Cheng, Zichen Ding, Liheng Chen, Paul~Pu Liang, and Yu~Qiao.
\newblock Os-atlas: A foundation action model for generalist gui agents, 2024.
\newblock URL \url{https://arxiv.org/abs/2410.23218}.

\bibitem[Xie et~al.(2025)Xie, Deng, Li, Yang, Wu, Chen, Hu, Wang, Xu, Wang, Xu, Wang, Sahoo, Yu, and Xiong]{xie2025scalingcomputerusegroundinguser}
Tianbao Xie, Jiaqi Deng, Xiaochuan Li, Junlin Yang, Haoyuan Wu, Jixuan Chen, Wenjing Hu, Xinyuan Wang, Yuhui Xu, Zekun Wang, Yiheng Xu, Junli Wang, Doyen Sahoo, Tao Yu, and Caiming Xiong.
\newblock Scaling computer-use grounding via user interface decomposition and synthesis, 2025.
\newblock URL \url{https://arxiv.org/abs/2505.13227}.

\bibitem[Yang et~al.(2024)Yang, Wang, Li, Luo, Chen, Huang, and Li]{yang2024ariauivisualgroundinggui}
Yuhao Yang, Yue Wang, Dongxu Li, Ziyang Luo, Bei Chen, Chao Huang, and Junnan Li.
\newblock Aria-ui: Visual grounding for gui instructions, 2024.
\newblock URL \url{https://arxiv.org/abs/2412.16256}.

\bibitem[Yuan et~al.(2025)Yuan, Zhang, Li, Cai, Yao, Chen, Wang, Hou, Chen, Jiang, and Li]{yuan2025enhancingvisualgroundinggui}
Xinbin Yuan, Jian Zhang, Kaixin Li, Zhuoxuan Cai, Lujian Yao, Jie Chen, Enguang Wang, Qibin Hou, Jinwei Chen, Peng-Tao Jiang, and Bo~Li.
\newblock Enhancing visual grounding for gui agents via self-evolutionary reinforcement learning.
\newblock 2025.
\newblock URL \url{https://arxiv.org/abs/2505.12370}.

\bibitem[Zhang et~al.(2024)Zhang, Li, He, Zhang, Qiao, Qin, Ma, Kang, Lin, Rajmohan, Zhang, and Zhang]{zhang2024ufouifocusedagentwindows}
Chaoyun Zhang, Liqun Li, Shilin He, Xu~Zhang, Bo~Qiao, Si~Qin, Minghua Ma, Yu~Kang, Qingwei Lin, Saravan Rajmohan, Dongmei Zhang, and Qi~Zhang.
\newblock Ufo: A ui-focused agent for windows os interaction, 2024.
\newblock URL \url{https://arxiv.org/abs/2402.07939}.

\bibitem[Zhang et~al.(2025{\natexlab{a}})Zhang, He, Qian, Li, Li, Qin, Kang, Ma, Liu, Lin, Rajmohan, Zhang, and Zhang]{zhang2025largelanguagemodelbrainedgui}
Chaoyun Zhang, Shilin He, Jiaxu Qian, Bowen Li, Liqun Li, Si~Qin, Yu~Kang, Minghua Ma, Guyue Liu, Qingwei Lin, Saravan Rajmohan, Dongmei Zhang, and Qi~Zhang.
\newblock Large language model-brained gui agents: A survey.
\newblock 2025{\natexlab{a}}.
\newblock URL \url{https://arxiv.org/abs/2411.18279}.

\bibitem[Zhang et~al.(2025{\natexlab{b}})Zhang, Huang, Ni, Mu, Qin, He, Wang, Yang, Zhao, Du, Li, Kang, Jiang, Zheng, Wang, Qian, Ma, Lou, Lin, Rajmohan, and Zhang]{zhang2025ufo2desktopagentos}
Chaoyun Zhang, He~Huang, Chiming Ni, Jian Mu, Si~Qin, Shilin He, Lu~Wang, Fangkai Yang, Pu~Zhao, Chao Du, Liqun Li, Yu~Kang, Zhao Jiang, Suzhen Zheng, Rujia Wang, Jiaxu Qian, Minghua Ma, Jian-Guang Lou, Qingwei Lin, Saravan Rajmohan, and Dongmei Zhang.
\newblock Ufo2: The desktop agentos.
\newblock 2025{\natexlab{b}}.
\newblock URL \url{https://arxiv.org/abs/2504.14603}.

\bibitem[Zhang et~al.(2023)Zhang, Yang, Liu, Han, Chen, Huang, Fu, and Yu]{zhang2023appagentmultimodalagentssmartphone}
Chi Zhang, Zhao Yang, Jiaxuan Liu, Yucheng Han, Xin Chen, Zebiao Huang, Bin Fu, and Gang Yu.
\newblock Appagent: Multimodal agents as smartphone users, 2023.
\newblock URL \url{https://arxiv.org/abs/2312.13771}.

\bibitem[Zhou et~al.(2025)Zhou, Dai, Wang, Zhou, Jia, and Xu]{zhou2025guig1understandingr1zeroliketraining}
Yuqi Zhou, Sunhao Dai, Shuai Wang, Kaiwen Zhou, Qinglin Jia, and Jun Xu.
\newblock Gui-g1: Understanding r1-zero-like training for visual grounding in gui agents.
\newblock 2025.
\newblock URL \url{https://arxiv.org/abs/2505.15810}.

\end{thebibliography}

\clearpage
\appendix
\section{Appendix}
\subsection{Analysis of Spurious Rewards}
Recent studies have shown that even spurious rewards can stimulate reinforcement learning training processes~\citep{shao2025spuriousrewardsrethinkingtraining}, raising important questions about reward design robustness. To better explore the impact of artificial reward signals on GUI grounding performance and validate the necessity of our proposed Point-to-Plane Gaussian reward mechanism, we conducted controlled experiments with two distinct fake reward strategies: 
(i) Random $U(0,1)$ Reward: rewards are randomly sampled from a uniform distribution $U(0,1)$ (including boundary values 0 and 1); (ii) Binary Random Reward: rewards are randomly assigned as either 0 or 1 with equal probability, creating maximum variance in sparse feedback patterns.

\begin{figure}[t]
    \centering
    \includegraphics[width=0.95\textwidth]{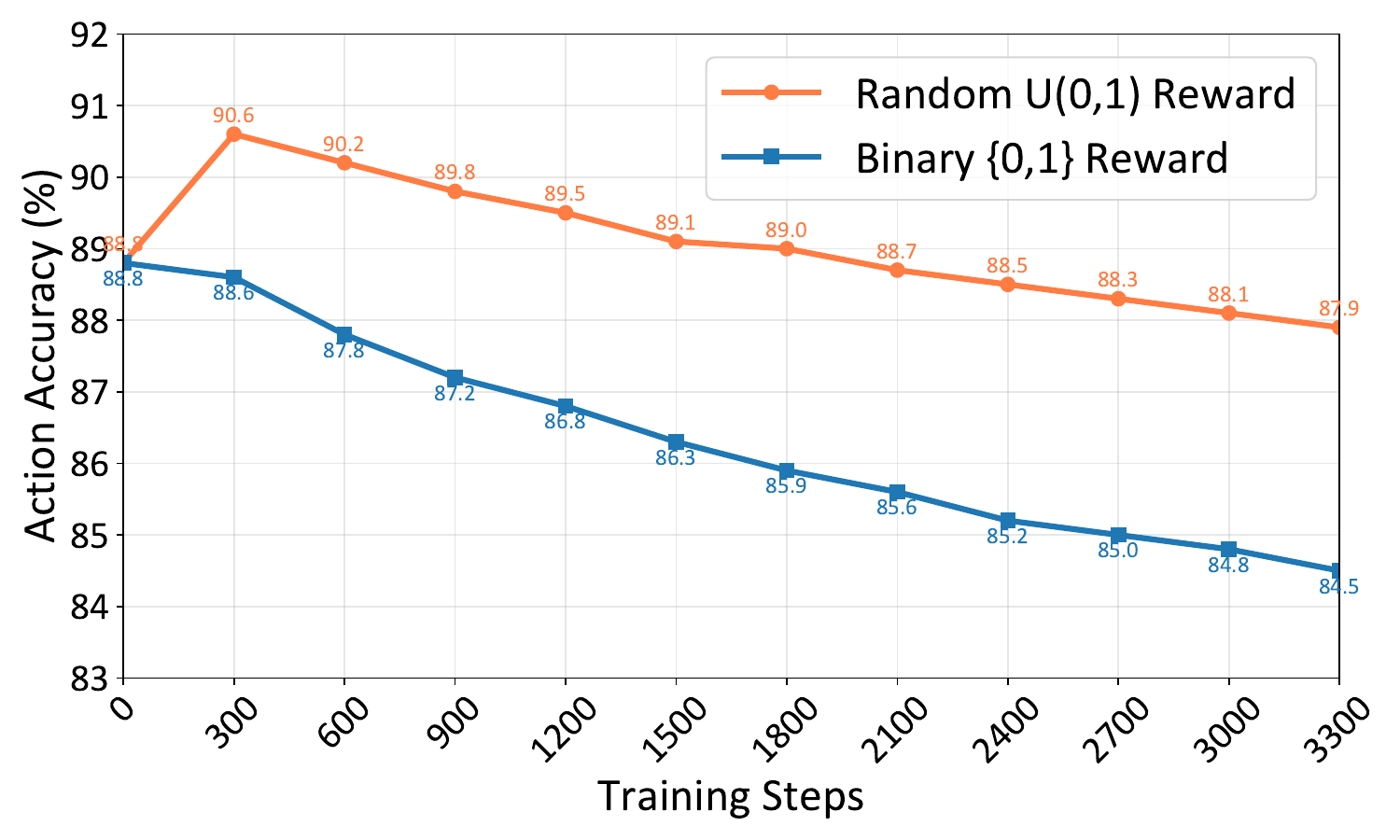}
    \caption{Performance comparison between random reward strategies on ScreenSpot-V2. Both continuous random $U(0,1)$ rewards and binary random rewards show progressive degradation, demonstrating that GUI grounding requires spatially-meaningful reward signals rather than arbitrary feedback.}
    \label{fig:fake_reward}
\end{figure}

\begin{figure}[htbp]
    \centering
    \includegraphics[width=1.0\textwidth]{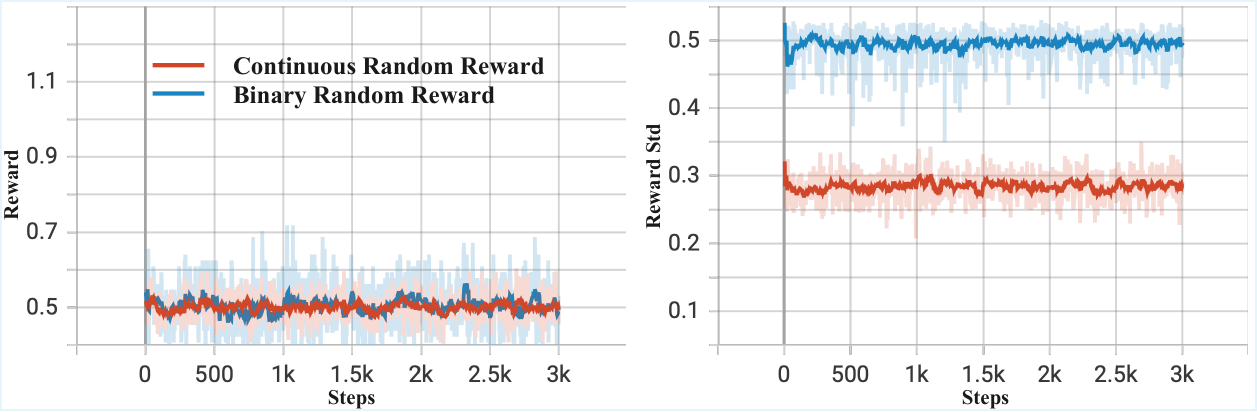}
    \caption{Reward distribution and standard deviation analysis during training. The reward variance patterns illustrate the fundamental differences between continuous and binary random reward mechanisms in reinforcement learning dynamics.}
    \label{fig:fake_reward_plot}
\end{figure}

As shown in Figure~\ref{fig:fake_reward} and~\ref{fig:fake_reward_plot}, our experimental results reveal three critical key findings: 
\textit{\textbf{(1) GUI Grounding Cannot Benefit Spurious Rewards for Effective Learning}}: 
Both random reward strategies exhibit progressive performance degradation with consistent downward trends. 
\begin{wrapfigure}[]{r}{0.5\textwidth}
\vspace{-0.8em}
\centering
\begin{tabular}{lc}
\toprule
\textbf{Hyperparameter} & \textbf{Value} \\
\midrule
    num\_generations & 8 \\
    per\_device\_train\_batch\_size & 8 \\
    gradient\_accumulation\_steps & 1 \\
    bf16 & true \\
    torch\_dtype & bfloat16 \\
    data\_seed & 42 \\
    gradient\_checkpointing & true \\
    attn\_implementation & flash\_attention\_2 \\
    num\_train\_epochs & 1 \\
    max\_pixels & 12845056 \\
    $\beta$ & 0.04 \\
    $\alpha$ & 0.5 \\
    $\nu$ & 1.0 \\
    $\gamma$ & 1.0 \\
\bottomrule
\end{tabular}
\caption{Hyperparameter settings used in the training experiments.}
\label{tab:hyperparameters}
\vspace{-1.2em}
\end{wrapfigure}
Continuous random rewards decline from 90.6\% to 87.9\% (-2.7\%) and binary random rewards drop from 88.6\% to 84.5\% (-4.1\%) over 3000 steps. These spurious rewards fail to provide effective learning signals for GUI grounding tasks, leading to gradual performance deterioration. Unlike other domains where random rewards may provide training benefits, GUI grounding tasks cannot benefit from arbitrary reward signals, demonstrating that meaningful spatial feedback is essential for effective learning. The failure of spurious rewards validates the effectiveness of our Point-to-Plane Gaussian reward mechanism. \textit{\textbf{(2) Continuous Random Rewards Show Superior Initial Performance}}: The $U(0,1)$ strategy maintains higher initial accuracy (90.6\% vs 88.6\%) with more gradual degradation. This difference stems from the fundamental learning signal availability: continuous random rewards consistently provide non-zero feedback at every training step, ensuring gradient flow and parameter updates throughout the learning process. In contrast, the binary strategy introduces complete signal absence (zero rewards) with 50\% probability, creating intermittent learning interruptions. During early training phases, these zero rewards introduce excessive noise that immediately disrupts gradient estimation and blocks policy updates, causing faster knowledge degradation. The continuous feedback mechanism, despite being random, maintains smoother gradient dynamics compared to the sporadic learning signals in binary rewards, highlighting the critical importance of consistent reward availability in reinforcement learning systems.

\tcbset{
  myboxstyle/.style={
    colback=gray!20,     
    colframe=black!70,    
    coltitle=white,       
    fonttitle=\bfseries,  
     fontupper=\itshape,
    boxrule=0.8mm,       
    arc=1mm,               
    boxsep=1mm,             
    left=1mm,              
    right=1mm,              
    top=1mm,               
    bottom=1mm,             
    toptitle=0mm,           
    bottomtitle=0mm,     
    enhanced,                  
  }
}



\subsection{Evaluation Details}
\label{sec:training_details}
\textbf{Compared Methods.} To better evaluate the advantages of our P2G reward mechanism, we assess existing methods that employ the RL paradigm for training as follows:

\begin{itemize}
\item \textbf{UI-R1}~\citep{lu2025uir1enhancingefficientaction}: Employs traditional sparse point rewards, assigning a reward of 1 when predicted coordinates [x,y] fall within the ground truth bounding box, and 0 otherwise.

\item \textbf{GUI-R1}~\citep{luo2025guir1generalistr1style}: Adopts the same sparse point reward strategy as UI-R1, specifically designed as a binary reward mechanism for GUI Grounding tasks.

\item \textbf{GUI-G1}~\citep{zhou2025guig1understandingr1zeroliketraining}: Combines sparse point rewards with IoU rewards for joint optimization, and introduces an adaptive reward function based on predicted bounding box size to handle GUI elements of different scales.

\item \textbf{InfiGUI-R1}~\citep{liu2025infiguir1advancingmultimodalgui}: Simultaneously utilizes sparse point rewards and sparse IoU rewards for optimization, enhancing GUI element localization accuracy through dual sparse reward mechanisms.

\item \textbf{SE-GUI}~\citep{yuan2025enhancingvisualgroundinggui}: Adopts a continuous reward function based on normalized distance, providing different reward values according to whether the predicted point is within the target bounding box and its distance from the center point.

\item \textbf{LPO}~\citep{tang2025lpoaccurateguiagent}: Implements a dynamic location reward mechanism that provides continuous reward feedback based on spatial accuracy by calculating the Euclidean distance between executed coordinates and target coordinates.
\end{itemize}

This comprehensive comparison encompasses diverse reward paradigms ranging from sparse binary mechanisms to continuous distance-based formulations, enabling thorough validation of our proposed \methodname approach across different methodological frameworks.

\begin{tcolorbox}[
    title=Thinking Prompt,
    label=prompt_thinking
]
\texttt{\{problem\}} 
Output the thinking process in $<$think$>$ $<$/think$>$ and final answer in $<$answer$>$ [x1,y1,x2,y2] $<$/answer$>$ tags.
\end{tcolorbox}

\begin{tcolorbox}[
    title=No Thinking Prompt,
    label=prompt_no_thinking
]
Outline the position corresponding to the instruction: \texttt{\{problem\}}. The output should be only [x1,y1,x2,y2].
\end{tcolorbox}

\subsection{Error Analysis}

\begin{figure}[t]
    \centering
    \begin{minipage}[b]{0.49\textwidth}
        \centering
        \includegraphics[width=\textwidth]{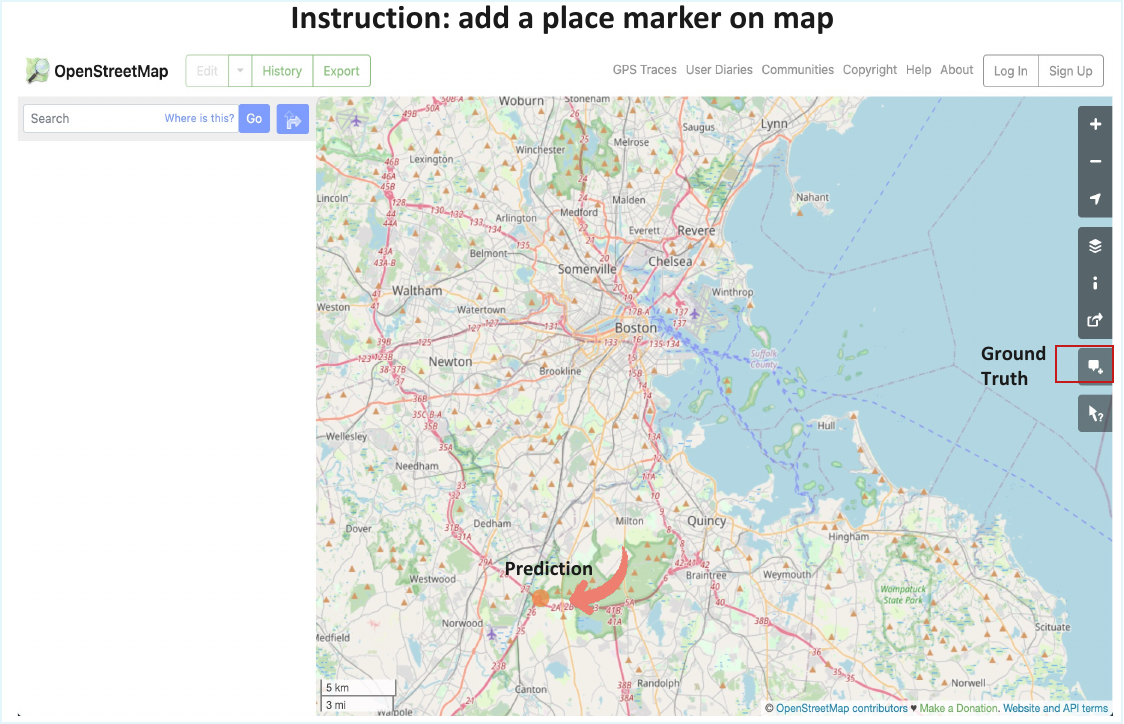}
        \subcaption{Information-Dense Interface}
        \label{figure:error_a}
    \end{minipage}
    \hfill
    \begin{minipage}[b]{0.49\textwidth}
        \centering
        \includegraphics[width=\textwidth]{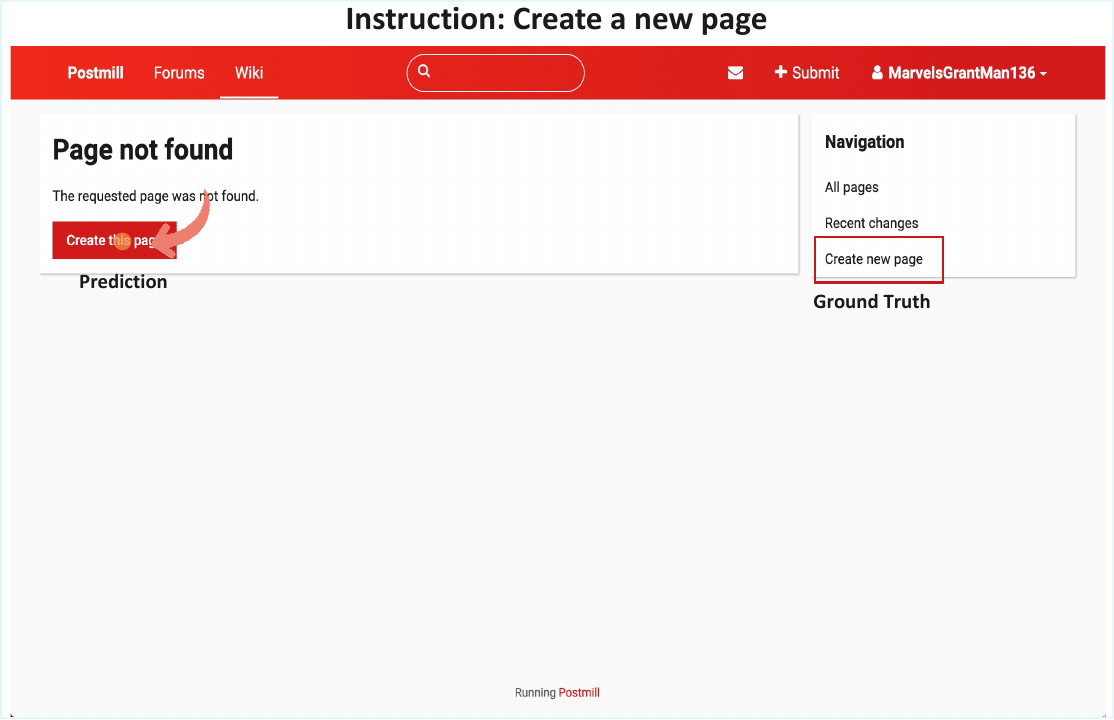}
        \subcaption{Visual and Structural Ambiguity}
        \label{figure:error_b}
    \end{minipage}    
    \caption{Analysis of GUI grounding failure cases.
    }
    \label{fig:error_analysis}
\end{figure}
We analyzed failure cases from the ScreenSpot-V2 dataset and found that icon recognition remains a critical challenge in GUI grounding. Our analysis of error distribution reveals that icon errors account for 76.9\% of total failures (63 icon errors vs 19 text errors across all platforms). This stark disparity underscores that semantic interpretation of visual symbols represents a fundamental bottleneck, as icons require models to infer abstract functionality from visual representations rather than explicit textual information. Beyond icon challenges, we identified two primary failure patterns: information-dense interface bottlenecks, where environments with high information density such as online maps or complex software interfaces (Figure~\ref{figure:error_a}) overwhelm the model's processing capabilities due to overlapping elements and complex visual hierarchies; and visual and structural ambiguity, where the model becomes confused when multiple UI elements share similar visual features or spatial arrangements (Figure~\ref{figure:error_b}), leading to incorrect selections among viable candidates when task descriptions lack sufficient specificity. These findings highlight that current GUI grounding limitations stem primarily from semantic understanding challenges rather than spatial localization capabilities, suggesting that future research should prioritize enhanced visual-semantic reasoning to bridge the gap between visual perception and functional intent.

\subsection{Future Work}
As GUI agents scale to handle complex high-resolution interfaces, computational overhead may become a limiting factor for practical deployment. Future work could explore model compression techniques~\citep{liu2025shiftingaiefficiencymodelcentric} and acceleration frameworks for large vision-language models~\citep{liu2025videocompressioncommanderplugandplay,liu2025globalcompressioncommanderplugandplay} to reduce inference costs while preserving grounding performance. Such optimizations would enable broader adoption of GUI agents across diverse computing environments.

\end{document}